\documentclass[leqno]{amsart}

\usepackage[foot]{amsaddr}

\usepackage{color}
\usepackage{geometry}
\usepackage{subfiles}
\usepackage{graphicx}
\usepackage[section]{placeins}
\usepackage{hyperref}
\usepackage{stmaryrd}
\usepackage{amsmath,amssymb,amsfonts,amsthm,textcomp}
\usepackage{mathtools}
\usepackage{tensor}
\usepackage{multicol}
\usepackage{subfig}
\usepackage{array}
\usepackage{accents}
\usepackage[colorinlistoftodos]{todonotes}
\usepackage[ruled,vlined]{algorithm2e}
\graphicspath{
              {./figures/}
             }

\newfont{\tenbfsl}{cmbxti9 scaled 1200}
\newfont{\tenbbb}{msbm10}
\newfont{\svnbbb}{msbm8}

\newcommand{\bs}[1]{\boldsymbol{#1}}
\newcommand{\cl}[1]{\mathcal{#1}}

\newcommand{\bb}[1]{\mathbb{#1}}
\newcommand{\norm}[2]{\left\lVert {#1} \right\rVert_{#2}}

\newcommand{\fr}[2]{\textstyle{\frac{{#1}}{{#2}}}}

\newcommand{\dv}{\,\mathrm{d}v}
\newcommand{\da}{\,\mathrm{d}a}

\newcommand{\Grad}{\mathrm{grad}\mskip2mu}
\newcommand{\Div}{\mathrm{div}\mskip2mu}
\newcommand{\Curl}{\mathrm{curl}\mskip2mu}

\newcommand{\bsuI}{\bs{\upsilon}_{\scriptscriptstyle\cl{I}}}
\newcommand{\bssJ}{\bs{\varsigma}_{\scriptscriptstyle\cl{J}}}
\newcommand{\bsxA}{\bs{\xi}_{\scriptscriptstyle\cl{A}}}
\newcommand{\bszB}{\bs{\zeta}_{\scriptscriptstyle\cl{B}}}

\newcommand{\twovdots}{\mskip+2mu\colon\mskip-2mu}
\makeatletter
\def\threevdots{\mskip+4mu\vbox{\baselineskip2.25\p@ \lineskiplimit\z@
  \kern4.9\p@\hbox{.}\hbox{.}\hbox{.}}\mskip+3.8mu}
\makeatother

\newcommand{\PiSL}{\texttt{PiSL}}

\begin{document}

\title[Physics-informed Spectral Learning: the Discrete HHD]{Physics-informed Spectral Learning: the Discrete Helmholtz--Hodge Decomposition}
\author{Luis Espath$^{\sharp,1}$, Pouria Behnoudfar$^2$, \& Ra\'{u}l Tempone$^{3,4,5}$}
\address{$^1$School of Mathematical Sciences, University of Nottingham, Nottingham, NG7 2RD, United Kingdom}
\address{$^2$Mineral Resources, Commonwealth Scientific and Industrial Research Organisation (CSIRO), Kensington, Perth, WA 6152, Australia}
\address{$^3$Department of Mathematics, RWTH Aachen University, Geb\"{a}ude-1953 1.OG, Pontdriesch 14-16, 161, 52062 Aachen, Germany.}
\address{$^4$Alexander von Humboldt Professor in Mathematics for Uncertainty Quantification, RWTH Aachen University, Germany.}
\address{$^5$King Abdullah University of Science \& Technology (KAUST), Computer, Electrical and Mathematical Sciences \& Engineering Division (CEMSE), Thuwal 23955-6900, Saudi Arabia.}
\email{$^\sharp$espath@gmail.com}
\date{\today}

\begin{abstract}
\noindent
%auto-ignore
In this work, we further develop the Physics-informed Spectral Learning (\PiSL) by Espath et al. \cite{Esp21} based on a discrete $L^2$ projection to solve the discrete Hodge--Helmholtz decomposition from sparse data. Within this physics-informed statistical learning framework, we adaptively build a sparse set of Fourier basis functions with corresponding coefficients by solving a sequence of minimization problems where the set of basis functions is augmented greedily at each optimization problem. Moreover, our \PiSL\, computational framework enjoys spectral (exponential) convergence. We regularize the minimization problems with the seminorm of the fractional Sobolev space in a Tikhonov fashion. In the Fourier setting, the divergence- and curl-free constraints become a finite set of linear algebraic equations. The proposed computational framework combines supervised and unsupervised learning techniques in that we use data concomitantly with the projection onto divergence- and curl-free spaces. We assess the capabilities of our method in various numerical examples including the `Storm of the Century' with satellite data from 1993.
\\
\textbf{keywords:} Physics-informed Machine Learning; Sparse approximation; Statistical learning; Fluid mechanics.
\\
\textbf{AMS subject classifications:}
$\cdot$
76-10 % Mathematical modeling or simulation for problems pertaining to fluid mechanics
$\cdot$
68T05 %	Learning and adaptive systems in artificial intelligence
$\cdot$
30E10 % Approximation in the complex plane
$\cdot$
35L65 % Partial differential equations - Conservation laws
$\cdot$

\end{abstract}

\maketitle

\tableofcontents                        % Print table of contents

%-------------------------------------------------------------------------------%

%auto-ignore
\section{Introduction}

Machine learning is an essential and universal tool for handling and understanding data. Nonetheless, machine learning capabilities are not restricted to generating data-driven models. Various fluid flow applications have been extensively developed in recent years. Particular attention has been focused on Physics-informed Neural Networks (\text{PiNNs}) \cite{Ram19,Maz20} in a statistical learning context, that is, data- and model-driven machine learning models. Such models combine measurements with underlying physical laws to improve the reconstruction quality, especially when data are sparse. A comprehensive review of machine learning for fluid mechanics is presented in \cite{Bru20}. Nonetheless, incorporating physical constraints along with other mathematical models instead of neural networks to achieve good reconstruction results has been applied before the machine learning era.

This work presents further developments of Physics-informed Spectral Learning (\PiSL) by Espath et al. \cite{Esp21} based on a discrete $L^2$ projection to solve the discrete Helmholtz--Hodge Decomposition (HDD) from sparse data. Given a finite set of vector measurements, we aim to carry out HHD. Our \PiSL\ learning technique is a physically informed statistical learning framework. Our approximation is based on Fourier basis functions and is sparse. In an adaptive manner, we start with a minimal set of Fourier basis functions and construct a more extensive sparse basis function set along a sequence of optimization problems. We increase the basis function set in each optimization problem and measure the relative energy. We retain only the most energetic high-wavenumber modes, and the optimization sequence continues. These optimization problems are ill-posed; thus, we use a Tikhonov regularizer based on the seminorm of the fractional Sobolev space. Additionally, our \PiSL\ approximation uses the properties of the Fourier series to construct pointwise divergence- and curl-free fields that are orthogonal in the $L^2$ sense.

The discrete HHD has numerous applications in fluid flows, electromagnetism, and image processing problems. We recall two applications to image processing problems: hurricane tracking and fingerprint analysis, studied by Palit et al. \cite{Pal05}. The HHD was used to identify the center of rotation, specifically the hurricane eye. The HHD was also used to identify fingerprint reference points, namely the point with the maximum curvature.

The remainder of this work is organized as follows. The mathematical notation and problem statement are presented in \S\ref{sc:notation.problem}, and the \PiSL\ method for spatial data is introduced in \S\ref{sc:PiSL}. The accuracy of this method is assessed in numerical examples in \S\ref{sc:numerics}. Conclusions are drawn in \S\ref{sc:conclusions}.

%auto-ignore
\section{Notation and problem statement}
\label{sc:notation.problem}

We let $\bs{\xi}\colon\cl{D}\subset\bb{R}^n\mapsto\bb{R}^n$ denote a real-valued vector field, where $\cl{D}\coloneqq\Pi_{\iota=1}^n[0, D_\iota]$ is a physical domain of length $D_\iota$ per direction $\iota$. We consider a set $\{\bs{u}_i\}_{i=1}^P$ of observed pointwise vector measurements of an unknown vector field at $\{\bs{x}_i\}_{i=1}^P\subset\cl{D}$. We aim to reconstruct an unknown vector field from the measurements $\{\bs{u}_i\}_{i=1}^P$ while performing HHD. We also aim to compute the divergence- and curl-free components of the underlying unknown vector field.

We construct our approximation on the fractional Sobolev space of all periodic functions that are square integrable on a toroidal $\cl{D}$. Thus, by setting
\begin{equation}\label{eq:alpha}
\hat{\bs{\alpha}}=(\alpha_1/D_1,\ldots,\alpha_n/D_n)\qquad\text{with}\qquad\bs{\alpha}\in\bb{Z}^n,
\end{equation}
where $k\in(1,\infty)$, we respectively define the $L^2$ and $H^k\coloneqq{W}^{k,2}$ spaces of vector-valued functions $\bs{\xi}\colon\cl{D}\mapsto\bb{R}^n$ as
\begin{equation}\label{eq:L2}
L^2(\cl{D})\coloneqq\left\{\bs{\xi}=\sum\limits_{\bs{\alpha} \in \bb{Z}^n} \bs{\xi}_{\bs{\alpha}} \exp(2\pi\jmath\,\hat{\bs{\alpha}}\cdot\bs{x})\,\bigg\vert\,\dfrac{1}{|\cl{D}|}\int_\cl{D}\|\bs{\xi}\|^2\dv=\sum\limits_{\bs{\alpha} \in \bb{Z}^n}\|\bs{\xi}_{\bs{\alpha}}\|^2<\infty\right\},
\end{equation}
and
\begin{equation}\label{eq:Hp}
H^{k}(\cl{D})\coloneqq\left\{\bs{\xi}\in L^2(\cl{D})\,\bigg\vert\,\sum_{\bs{\alpha}\in\bb{Z}^n}(2\pi\|\hat{\bs{\alpha}}\|)^{2k}\|\bs{\xi}_{\bs{\alpha}}\|^2<\infty\right\},
\end{equation}
where $\jmath\coloneqq\sqrt{-1}$ is the imaginary unit and $\bs{\xi}_{\bs{\alpha}}$ represents the $\bs{\alpha}$th component of the Fourier transform of $\bs{\xi}$, that is,
\begin{equation}
\bs{\xi}_{\bs{\alpha}}\coloneqq\dfrac{1}{|\cl{D}|}\int_{\cl{D}}\bs{\xi}(\bs{x})\exp(-2\pi\jmath\mskip2mu\hat{\bs{\alpha}}\cdot\bs{x})\dv\qquad\forall\,\bs{\alpha}\in\bb{Z}^n.
\end{equation}
In expression \eqref{eq:Hp}, we define the seminorm of $H^k$ for the vector-valued $\bs{\xi}$. However, the norm that induces the space $H^k$ is
\begin{equation}
\norm{\bs{\xi}}{H^k(\cl{D})}^2\coloneqq\sum_{\bs{\alpha}\in\bb{Z}^n}(1+(2\pi\|\hat{\bs{\alpha}}\|)^{2k})\|\bs{\xi}_{\bs{\alpha}}\|^2.
\end{equation}
In what follows, $\Grad^k\bs{\xi}$ is the $k$th gradient of $\bs{\xi}$ and $\mathring{H}^k$ denotes the fractional seminorm
\begin{equation}\label{eq:regularization}
\norm{\bs{\xi}}{\mathring{H}^k(\cl{D})}^2\coloneqq\sum_{\bs{\alpha}\in\bb{Z}^n}(2\pi\|\hat{\bs{\alpha}}\|)^{2k}\|\bs{\xi}_{\bs{\alpha}}\|^2=\norm{\Grad^k\bs{\xi}}{L^2(\cl{D})}^2.
\end{equation}
Alternatively, the $L^2$ inner product for vector-valued functions may assume the following conventional form in an arbitrary domain, $\cl{D}$, and on its boundary, $\partial\cl{D}$,
\begin{equation}
(\bs{\xi},\bs{\omega})_{L^2(\cl{D})}\coloneqq\dfrac{1}{|\cl{D}|}\int_{\cl{D}}\bs{\xi}\cdot\bs{\omega}^\ast\dv\qquad\text{and}\qquad(\bs{\xi},\bs{\omega})_{L^2(\partial\cl{D})}\coloneqq\dfrac{1}{|\partial\cl{D}|}\int_{\partial\cl{D}}\bs{\xi}\cdot\bs{\omega}^\ast\da,
\end{equation}
where $\bs{\xi}$ and $\bs{\omega}$ are vector fields, and the asterisk denotes the complex-conjugate pair.

We can now state the discrete Helmholtz--Hodge decomposition problem as follows. Given $k_{\mathrm{d}}>1$, $k_{\mathrm{c}}>1$, $\epsilon_{\mathrm{d}}>0$, and $\epsilon_{\mathrm{c}}>0$, find $(\bs{\upsilon}^{\mathrm{opt}},\bs{\varsigma}^{\mathrm{opt}})$ such that
\begin{equation}\label{eq:optimization}
\left\{
\begin{aligned}
&(\bs{\upsilon}^{\mathrm{opt}},\bs{\varsigma}^{\mathrm{opt}}) \coloneqq \underset{\substack{\bs{\upsilon}\in H^{k_\mathrm{d}}(\cl{D}) \\ \bs{\varsigma}\in H^{k_\mathrm{c}}(\cl{D})}}{\arg\min} \, \dfrac{1}{P}\sum_{i=1}^P\norm{\bs{\upsilon}({\bs{x}}_i) + \bs{\varsigma}({\bs{x}}_i) - {\bs{u}_i}}{}^2 + \epsilon_{\mathrm{d}}\norm{\bs{\upsilon}}{\mathring{H}^{k_{\mathrm{d}}}(\cl{D})}^2 + \epsilon_{\mathrm{c}}\norm{\bs{\varsigma}}{\mathring{H}^{k_{\mathrm{c}}}(\cl{D})}^2,\\[4pt]
&\text{subject to } \Div\bs{\upsilon}=0 \,\wedge\, \Curl\bs{\varsigma}=\bs{0},
\end{aligned}
\right.
\end{equation}
where $\Div\bs{\upsilon}$ and $\Curl\bs{\varsigma}$ represent the divergence of $\bs{\upsilon}$ and the curl of $\bs{\varsigma}$, respectively. The parameters $\epsilon_{\mathrm{d}}$ and $\epsilon_{\mathrm{c}}$ are regularization parameters for the divergence- and curl-free fields, respectively. The third term in \eqref{eq:optimization} is a regularization term that penalizes the seminorm $\mathring{H}^k$ of the fields, where $k$ is either $k_{\mathrm{d}}$ or $k_{\mathrm{c}}$.

From the embedding Sobolev theorem (see the Appendix in Espath et al. \cite{Esp21}) with $k>n/p$, we have that $\bs{\upsilon}$ and $\bs{\varsigma}$ belong to the H\"{o}lder space ${C}^{k-\left[\frac{n}{p}\right]-1,\gamma}(\cl{D})$, namely, H\"{o}lder continuous with some positive exponent $\gamma$. Thus, for two-dimensional problems $n=2$ and $p=2$ (for $L^{p=2}$ spaces), we have that $k>1$ whereas in three-dimensional problems $n=3$ and $p=2$, we have $k>1.5$.

\section{Spatial approximation: Physics-informed Spectral Learning (\texttt{PiSL})}
\label{sc:PiSL}

Next, consider the following finite-dimensional representation $\bsuI$ and $\bssJ$ in $H^k(\cl{D})$ as follows.
\begin{equation}\label{eq:fourier.approximation}
\begin{aligned}
\bsuI(\bs{x}) \coloneqq \sum\limits_{\bs{\alpha} \in \cl{I}} \bs{\upsilon}_{\bs{\alpha}} \varphi(\bs{x},\bs{\alpha}) \qquad\text{and}\qquad \bssJ(\bs{x}) \coloneqq \sum\limits_{\bs{\alpha} \in \cl{J}} \bs{\varsigma}_{\bs{\alpha}} \varphi(\bs{x},\bs{\alpha}),\\[4pt]
\text{where}\qquad\varphi(\bs{x},\bs{\alpha}) \coloneqq \exp(2\pi\jmath\,\hat{\bs{\alpha}}\cdot\bs{x})\qquad\forall\,\bs{x}\in\Pi_{\beta=1}^n[0,D_\beta]\wedge\bs{\alpha}\in\cl{I},\cl{J},
\end{aligned}
\end{equation}
where $\cl{I}$ and $\cl{J}\subset\bb{Z}^n$ are finite index sets of tuples comprising $n$ integers defining the indices of the basis functions and $\bs{\upsilon}_{\bs{\alpha}},\,\bs{\varsigma}_{\bs{\alpha}}\,\in\bb{C}^n$ are their Fourier coefficients for all $\bs{\alpha}\in\cl{I}$ and $\bs{\alpha}\in\cl{J}$, respectively.

\subsection{Real value, divergence-, and curl-free constraints}

Here, we impose the fields $\bsuI$ and $\bssJ$ as divergence- and curl-free fields, respectively. For this, the divergence and curl constraints become linear algebraic constraints by the differentiability properties of trigonometric functions.

First, the gradient of the basis function $\varphi$ with respect to the spatial coordinate $\bs{x}$ is
\begin{equation}
\Grad\varphi(\bs{x},\bs{\alpha}) = 2 \pi \jmath \varphi(\bs{x},\bs{\alpha}) \mskip3mu \hat{\bs{\alpha}}.
\end{equation}
Then, we can write the divergence and curl as follows:
\begin{equation}
\Div\bsuI = 2 \pi \jmath \sum\limits_{\bs{\alpha} \in \cl{I}} (\bs{\upsilon}_{\bs{\alpha}} \cdot \hat{\bs{\alpha}}) \mskip3mu \varphi(\bs{x},\bs{\alpha}), \qquad\text{and}\qquad \Curl\bssJ = 2 \pi \jmath \sum\limits_{\bs{\alpha} \in \cl{J}} (\bs{\varsigma}_{\bs{\alpha}} \times \hat{\bs{\alpha}}) \mskip3mu \varphi(\bs{x},\bs{\alpha}).
\end{equation}
Thus, from
\begin{equation}
\Div\bsuI=0, \qquad\text{and}\qquad \Curl\bssJ=\bs{0},
\end{equation}
we respectively arrive at
\begin{equation}\label{eq:algebraic.div.constraint}
\hat{\bs{\alpha}}\cdot\bs{\upsilon}_{\bs{\alpha}} = 0 \qquad\forall\,\bs{\alpha}\in\cl{I}, \qquad\text{and}\qquad \hat{\bs{\alpha}}\times\bs{\varsigma}_{\bs{\alpha}} = \bs{0} \qquad\forall\,\bs{\alpha}\in\cl{J}.
\end{equation}
In addition, \eqref{eq:algebraic.div.constraint}$_2$ implies that
\begin{equation}\label{eq:algebraic.div.constraint.consequence}
\hat{\bs{\alpha}}\times\bs{\varsigma}_{-\bs{\alpha}} = \bs{0} \qquad\forall\,\bs{\alpha}\in\cl{J}.
\end{equation}

To obtain real-valued representations for $\bsuI$ and $\bssJ$, we impose additional algebraic constraints:
\begin{equation}\label{eq:algebraic.real.constraint}
\bs{\upsilon}_{\bs{\alpha}}=\bs{\upsilon}_{-\bs{\alpha}}^\ast\qquad\forall\,\bs{\alpha}\in\cl{I}, \qquad\text{and}\qquad \bs{\varsigma}_{\bs{\alpha}}=\bs{\varsigma}_{-\bs{\alpha}}^\ast\qquad\forall\,\bs{\alpha}\in\cl{J}.
\end{equation}
Lastly, in view of \eqref{eq:algebraic.div.constraint} and \eqref{eq:algebraic.real.constraint}, we also arrive at the following pointwise orthogonality conditions
\begin{equation}\label{eq:orthogonality.coefficients.1}
\bs{\upsilon}_{\bs{\alpha}} \cdot \bs{\varsigma}_{\bs{\alpha}} = 0, \qquad \bs{\upsilon}_{\bs{\alpha}} \cdot \bs{\varsigma}^\ast_{-\bs{\alpha}} = 0 \qquad \bs{\upsilon}^\ast_{-\bs{\alpha}} \cdot \bs{\varsigma}_{\bs{\alpha}} = 0, \qquad \bs{\upsilon}^\ast_{-\bs{\alpha}} \cdot \bs{\varsigma}^\ast_{-\bs{\alpha}} = 0,
\end{equation}
while, in view of \eqref{eq:algebraic.div.constraint.consequence} and \eqref{eq:algebraic.real.constraint}, we have that
\begin{equation}\label{eq:orthogonality.coefficients.2}
\bs{\upsilon}_{\bs{\alpha}} \cdot \bs{\varsigma}_{-\bs{\alpha}} = 0, \qquad \bs{\upsilon}_{\bs{\alpha}} \cdot \bs{\varsigma}^\ast_{\bs{\alpha}} = 0, \qquad \bs{\upsilon}^\ast_{-\bs{\alpha}} \cdot \bs{\varsigma}_{-\bs{\alpha}} = 0, \qquad\text{and}\qquad  \bs{\upsilon}^\ast_{-\bs{\alpha}} \cdot \bs{\varsigma}^\ast_{\bs{\alpha}} = 0 \qquad \forall \bs{\alpha} \in \cl{I}\cap\cl{J}.
\end{equation}

\subsection{$L^2$ orthogonality}

Granted that $\bsuI$ and $\bssJ$ are orthogonal in the $L^2$ sense, and $\bsuI\cdot\bs{n} = 0$ on $\partial\cl{D}$, where $\bs{n}$ is the unit outward normal to $\partial\cl{D}$, the Helmholtz--Hodge decomposition is unique, see for instance, the book by Chorin \cite{Cho90}. We now demonstrate that our sparse spectral approximations are orthogonal in the $L^2$ sense by construction.

Next, consider the $L^2$ inner product
\begin{align}
(\bsuI,\bssJ)_{L^2(\cl{D})} & \coloneqq \int\limits_{\cl{D}} \bsuI \cdot \bssJ^\ast \dv = \bigg(\sum_{\bs{\alpha} \in \cl{I}} \bs{\upsilon}_{\bs{\alpha}} \varphi(\bs{x},\bs{\alpha}), \sum_{\bs{\beta} \in \cl{J}} \bs{\varsigma}^\ast_{\bs{\beta}} \varphi^\ast(\bs{x},\bs{\beta})\bigg)_{L^2(\cl{D})}, \nonumber \\[4pt]
& = \sum_{\bs{\alpha} \in \cl{I} \cap \cl{J}} \bs{\upsilon}_{\bs{\alpha}} \cdot \bs{\varsigma}^\ast_{\bs{\alpha}}.\footnotemark
\end{align}
\footnotetext{See Appendix \ref{sc:inner.product}, Equation \eqref{eq:parseval.L2}.}Additionally, considering the expression \eqref{eq:orthogonality.coefficients.2}$_2$, we obtain $\bs{\upsilon}_{\bs{\alpha}} \cdot \bs{\varsigma}^\ast_{\bs{\alpha}} = 0$. Thus
\begin{equation}
(\bsuI,\bssJ)_{L^2(\cl{D})} = 0.
\end{equation}
Therefore, $\bsuI$ and $\bssJ$ are orthogonal in the $L^2$ sense.

\subsection{Reformulation}

In the finite Fourier representation \eqref{eq:fourier.approximation} augmented by the algebraic constraints \eqref{eq:algebraic.div.constraint} and \eqref{eq:algebraic.real.constraint} over $\cl{I}$ and $\cl{J}$, the optimization problem \eqref{eq:optimization}, where
\begin{equation}\label{eq:f}
\cl{F}(\bsuI,\bssJ) \coloneqq \dfrac{1}{P}\sum_{i=1}^P\norm{\bsuI({\bs{x}}_i) + \bssJ({\bs{x}}_i) - {\bs{u}_i}}{}^2 + \epsilon_{\mathrm{d}}\norm{\bsuI}{\mathring{H}_{\mathrm{div}}^{k_{\mathrm{d}}}(\cl{D})}^2 + \epsilon_{\mathrm{c}}\norm{\bssJ}{\mathring{H}_{\mathrm{curl}}^{k_{\mathrm{c}}}(\cl{D})}^2,
\end{equation}
becomes
\begin{equation}\label{eq:reformulated.optimization}
(\bsuI^{\mathrm{opt}},\bssJ^{\mathrm{opt}})\coloneqq\underset{\substack{\bsuI\in H^{k_\mathrm{d}}_{\mathrm{div}}(\cl{D}) \\ \bssJ\in H^{k_\mathrm{c}}_{\mathrm{curl}}(\cl{D})}}{\arg\min} \, \cl{F}(\bsuI,\bssJ),
\end{equation}
where
\begin{equation}
H^{k_\mathrm{d}}_{\mathrm{div}}(\cl{D}) \coloneqq \{\bsuI \in H^{k_\mathrm{d}}(\cl{D}) | \hat{\bs{\alpha}}\cdot\bs{\upsilon}_{\bs{\alpha}} = 0 \,\wedge\,\bs{\upsilon}_{\bs{\alpha}}-\bs{\upsilon}_{-\bs{\alpha}}^\ast=\bs{0}, \,\forall\,\bs{\alpha}\in\cl{I}\},
\end{equation}
and
\begin{equation}
H^{k_\mathrm{c}}_{\mathrm{curl}}(\cl{D}) \coloneqq \{\bssJ \in H^{k_\mathrm{c}}(\cl{D}) | \hat{\bs{\alpha}}\times\bs{\varsigma}_{\bs{\alpha}} = \bs{0} \,\wedge\,\bs{\varsigma}_{\bs{\alpha}}-\bs{\varsigma}_{-\bs{\alpha}}^\ast=\bs{0}, \,\forall\,\bs{\alpha}\in\cl{J}\}.
\end{equation}

Lastly, note that for a generic $\bsxA$ \PiSL\ approximation, as the bases are orthogonal, the regularization term in \eqref{eq:optimization}, defined in \eqref{eq:regularization} for $k\in\bb{R}$, for a generic field $\bsxA$ reads
\begin{align}
\norm{\bsxA}{\mathring{H}^k(\cl{D})}^2 & = \dfrac{1}{|\cl{D}|}\int_{\cl{D}} \|\Grad^k\bsxA\|^2 \dv = \dfrac{1}{|\cl{D}|}\int_{\cl{D}} \Grad^k\bsxA \cdot (\Grad^k\bsxA)^\ast \dv \nonumber\\
&= (2\pi)^{2k}\sum\limits_{\bs{\alpha} \in \cl{A}} \left(\bs{\xi}_{\bs{\alpha}}\cdot\bs{\xi}_{\bs{\alpha}}^\ast \right) (\hat{\bs{\alpha}}\cdot\hat{\bs{\alpha}})^k.\footnotemark
\end{align}
\footnotetext{See Appendix \ref{sc:inner.product}, Equation \eqref{eq:parseval.Hk}.}Moreover, given the spectral properties of Fourier transforms, we expect to recover the spectral (exponential) convergence. That is, for a constant $q$
\begin{equation}
|\bs{\xi}_{\bs{\alpha}}| \sim \cl{O} (\mathrm{e}^{-q|\bs{\alpha}|^b}),
\end{equation}
where $b$ is the exponential index of convergence defined as
\begin{equation}
b \coloneqq \lim_{|\bs{\alpha}| \to \infty} \dfrac{\ln|\ln|\bs{\xi}_{\bs{\alpha}}||}{\ln|\bs{\alpha}|}.
\end{equation}
Such methods as Physics-informed Neural Networks and Finite Element Methods (see, for instance, the book by Strang \& Fix \cite{Str73}) are endowed with a poorer convergence rate, namely, a algebraic convergence rate. That is, for a constant $b$ and a discretization parameter $h$, the $L^2$ error is of order $\cl{O} (h^b)$. Here, $h$ may be the number of neurons in a Neural Network (with $b<0$) or the mesh size (with $b>0$) in the Finite Element method.

Lastly, to solve the minimization in \eqref{eq:reformulated.optimization}, one needs to compute the gradient of the objective function. To that end, the interested readers are referred to Appendix \ref{sc:grad} and Espath et al. \cite{Esp21}.

\subsection{Algorithm}

In view of the \PiSL\ approximation \eqref{eq:fourier.approximation} with $\cl{A}$ being a set of $n$-tuples such that $\bs{\alpha}\in\cl{A}$, let $\delta\bs{\alpha}^\iota$ be all possible $n$-tuples populated with zeros and ones. In two dimensions, the possible $\delta\bs{\alpha}^\iota$ are $\pm(1,0)$, $\pm(0,1)$, and $\pm(1,1)$. In three dimensions, the possible $\delta\bs{\alpha}^\iota$ are $\pm(1,0,0)$, $\pm(0,1,0)$, $\pm(0,0,1)$, $\pm(1,1,0)$, $\pm(1,0,1)$, $\pm(0,1,1)$, and $\pm(1,1,1)$. Next, let $\partial\cl{A}$ be the boundary of the index set $\cl{A}$, such that $\bs{\alpha}\in\partial\cl{A}$ if and only if $\bs{\alpha}+\delta\bs{\alpha}^\iota\not\in\cl{A}$ for some $\iota$. Whenever an element $\bs{\beta}$ is included in (excluded from) $\cl{A}$, the element $-\bs{\beta}$ must also be included in (excluded from) $\cl{A}$ to satisfy constraint \eqref{eq:algebraic.real.constraint}.

The energy of the \PiSL\ approximation is defined as
\begin{equation}
\varepsilon(\cl{A})\coloneqq\sum_{\bs{\alpha}\in\cl{A}} \bs{\xi}_{\bs{\alpha}} \cdot \bs{\xi}_{\bs{\alpha}}^\ast,
\end{equation}
and the boundary energy is
\begin{equation}
\varepsilon(\partial\cl{A})\coloneqq\sum_{\bs{\alpha}\in\partial\cl{A}} \bs{\xi}_{\bs{\alpha}} \cdot \bs{\xi}_{\bs{\alpha}}^\ast .
\end{equation}

We begin by solving \eqref{eq:reformulated.optimization} in the smallest hypercube index space $\cl{A}\coloneqq(-1,0,1)^n$. To arrive at $\cl{A}^1$, we augment the index space $\cl{A}^0$ to include $\bs{\alpha}+\delta\bs{\alpha}^\iota$ for all $\bs{\alpha}\in\partial\cl{A}^0$ and $1\le\iota\le3^n-1 \eqqcolon N$. To obtain a sparse approximation, we retain only a percentage of these indices on the boundary $\partial\cl{A}^1$. After solving \eqref{eq:reformulated.optimization} for $\cl{A}^1$, we remove the indices on $\partial\cl{A}^1$ that contribute less than a pre-established energy threshold, $\varepsilon_{\partial\cl{A}}$. This iterative procedure is considered to converge when the energy increase of the index-space augmentation is $\Delta\epsilon_{\partial\cl{A}}$ or lower. This procedure is performed independently for both approximations, $\bsuI$ and $\bssJ$. Algorithm~\ref{al:bases.augmentation} provides the details. More details about the sparse construction are found in \cite{Haj20}.

\begin{algorithm}[H]\label{al:bases.augmentation}
\SetAlgoLined
\KwResult{output: $\bsuI$}
 data: $\{\bs{u}_i\}_{i=1}^P$\;
 initialization: $\cl{I}\coloneqq(-1,0,1)^n$, $\cl{J}\coloneqq(-1,0,1)^n$, $\epsilon_{\mathrm{d}}$, $\epsilon_{\mathrm{c}}$, $k_{\mathrm{d}}$, $k_{\mathrm{c}}$, $\varepsilon_{\partial\cl{I}}$, $\varepsilon_{\partial\cl{J}}$, $\Delta\varepsilon_{\partial\cl{I}}$, $\Delta\varepsilon_{\partial\cl{J}}$, \texttt{total\_it}\;
 \While{\texttt{niter} $\le$ \texttt{total\_it}}{
 \If{\texttt{niter} $\neq 0$}{
  $\cl{I}\gets\cl{I}\cup\left\{\bigcup_{\bs{\alpha}\in\cl{I}}\{\bs{\alpha}+\delta\bs{\alpha}^\iota\}_{\iota=1}^N\right\}$\;
  $\cl{J}\gets\cl{J}\cup\left\{\bigcup_{\bs{\alpha}\in\cl{J}}\{\bs{\alpha}+\delta\bs{\alpha}^\iota\}_{\iota=1}^N\right\}$\;
  }
  get $\{\bs{\upsilon}_{\bs{\alpha}}\}_{\bs{\alpha}\in\cl{I}}$ and $\{\bs{\varsigma}_{\bs{\alpha}}\}_{\bs{\alpha}\in\cl{J}}$ from solving \eqref{eq:reformulated.optimization}\;
  \eIf{$\dfrac{\varepsilon(\partial\cl{I})}{\varepsilon(\cl{I})}>\Delta\varepsilon_{\partial\cl{I}}$ $\vee$ $\dfrac{\varepsilon(\partial\cl{J})}{\varepsilon(\cl{J})}>\Delta\varepsilon_{\partial\cl{J}}$}{
   remove $\bs{\alpha}\in\partial\cl{I}$ and $\bs{\alpha}\in\partial\cl{J}$ corresponding to the low energy components, maintaining a fraction $1-\varepsilon_{\partial\cl{I}}$ and $1-\varepsilon_{\partial\cl{J}}$, respectively, of the relative boundary energy\;
   }{
   return $\{\bs{\upsilon}_{\bs{\alpha}}\}_{\bs{\alpha}\in\cl{I}}$ and $\{\bs{\varsigma}_{\bs{\alpha}}\}_{\bs{\alpha}\in\cl{J}}$\;
  }
  $\texttt{niter} +=1$\;
 }
 \caption{\PiSL\ Physics-informed Spectral Learning algorithm}
\end{algorithm}

%auto-ignore
\section{Numerical experiments}
\label{sc:numerics}

In this section, we present some numerical experiments to assess the \PiSL's efficiency and accuracy. Important to the presentation of these results, to depict the velocity fields, is the use of arrowed streamlines where the thickness is proportional to the magnitude of the velocity.

When the exact field representation is known, we can compute the pointwise error as
\begin{equation}\label{eq:intensive.error}
e(\bs{x})\coloneqq\norm{\bs{\xi}(\bs{x})-\bsxA^{\mathrm{opt}}(\bs{x})}{},
\end{equation}
where $\bs{\xi}_{\cl{A}}^{\mathrm{opt}}$ is the \PiSL\ approximation. Analogously, we define the partwise error (continuous $L^2$) error as
\begin{equation}\label{eq:extensive.error}
E\coloneqq\norm{\bs{\upsilon}(\bs{\xi})-\bsxA^{\mathrm{opt}}(\bs{x})}{L^2(\cl{D})}=\left(\int\limits_{\cl{D}}e^2(\bs{x})\dv\right)^{1/2}.
\end{equation}

\FloatBarrier

\subsection{Two-dimensional \PiSL: counter-rotating vortices with sources and sinks}

We here consider the domain $\cl{D} = [D,D]$ with $D = 3\pi$. Now, to define the vector-valued function to be decomposed, consider the scalar-valued function
\begin{equation}\label{eq:scalar.function}
\psi (\bs{x};\bs{x}_0) = \exp(-\fr{1}{2} \|\bs{x}-\bs{x}_0\|^2),
\end{equation}
and the vector-valued functions
\begin{equation}\label{eq:analytical.field.construction}
\bs{\upsilon}_{\mathrm{d}} (\bs{x};\bs{x}_0) = - \partial_2 \psi (\bs{x};\bs{x}_0) \bs{e}_1 + \partial_1 \psi (\bs{x};\bs{x}_0) \bs{e}_2, \qquad \text{and} \qquad \bs{\upsilon}_{\mathrm{c}} (\bs{x};\bs{x}_0) = \Grad \psi (\bs{x};\bs{x}_0),
\end{equation}
where $\partial_i \coloneqq \partial / \partial x_i$ and $\bs{e}_i$ are the basis vectors. Also, note that the two-dimensional vector field $\bs{\upsilon}_{\mathrm{d}}$ is divergence-free while $\bs{\upsilon}_{\mathrm{c}}$ is curl-free.

Next, let $\bs{x}_1 = (1D/8,D/2)$, $\bs{x}_2 = (3D/8,D/2)$, $\bs{x}_3 = (5D/8,D/2)$, $\bs{x}_4 = (7D/8,D/2)$, $\bs{x}_5 = (-D/8,D/2)$, and $\bs{x}_6 = (9D/8,D/2)$. Lastly, consider the divergence-free field
\begin{equation}
\bs{u}_{\mathrm{d}}(\bs{x}) = - \bs{\upsilon}_{\mathrm{d}} (\bs{x};\bs{x}_1) + \bs{\upsilon}_{\mathrm{d}} (\bs{x};\bs{x}_2) - \bs{\upsilon}_{\mathrm{d}} (\bs{x};\bs{x}_3) + \bs{\upsilon}_{\mathrm{d}} (\bs{x};\bs{x}_4) + \bs{\upsilon}_{\mathrm{d}} (\bs{x};\bs{x}_5) - \bs{\upsilon}_{\mathrm{d}} (\bs{x};\bs{x}_6).
\end{equation}
Similarly, consider $\bs{x}_6 = (0,0)$, $\bs{x}_7 = (D,D)$, $\bs{x}_8 = (0,D)$, $\bs{x}_9 = (D,0)$, $\bs{x}_{10} = (D/2,D/4)$, $\bs{x}_{11} = (D/2,3D/4)$, $\bs{x}_{12} = (D/2,-D/4)$, $\bs{x}_{13} = (D/2,5D/4)$, and
\begin{equation}
\bs{u}_{\mathrm{c}}(\bs{x}) = \bs{\upsilon}_{\mathrm{c}} (\bs{x};\bs{x}_6) + \bs{\upsilon}_{\mathrm{c}} (\bs{x};\bs{x}_7) + \bs{\upsilon}_{\mathrm{c}} (\bs{x};\bs{x}_8) + \bs{\upsilon}_{\mathrm{c}} (\bs{x};\bs{x}_9) - \bs{\upsilon}_{\mathrm{c}} (\bs{x};\bs{x}_{10}) - \bs{\upsilon}_{\mathrm{c}} (\bs{x};\bs{x}_{11}) - \bs{\upsilon}_{\mathrm{c}} (\bs{x};\bs{x}_{12}) - \bs{\upsilon}_{\mathrm{c}} (\bs{x};\bs{x}_{13}).
\end{equation}
Then, the field we aim to reconstruct while performing the discrete $L^2$ HHD is given by
\begin{equation}
\bs{u}(\bs{x}) = \bs{u}_{\mathrm{d}}(\bs{x}) + \bs{u}_{\mathrm{c}}(\bs{x}).
\end{equation}

\begin{figure}[!h]
\centering
\subfloat[Index set: divergence-free field]{\label{fg:ex5.div.index}\qquad\qquad\qquad\includegraphics[]{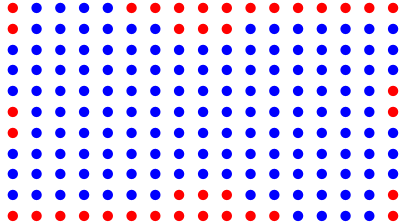}\qquad\qquad\qquad} \hspace{-2cm}
\subfloat[Index set: curl-free field]{\label{fg:ex5.curl.index}\qquad\qquad\qquad\includegraphics{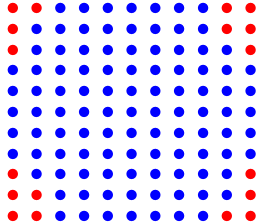}\qquad\qquad\qquad}
\caption{Panels \protect\subref{fg:ex5.div.index} and \protect\subref{fg:ex5.curl.index}: Index sets of the reconstructed \PiSL\, divergence- and curl-free fields, respectively. Blue dots represent the indices retained in the Fourier construction. Red dots represent indices removed from the Fourier construction.}
\label{fg:index.ex5}
\end{figure}

The discrete $L^2$ HHD is carried out using $250$ fixed measurements at random points in the domain $\cl{D}=[0,3\pi]^2$. We set the residual boundary energy $\varepsilon_{\partial\cl{I}}$ and $\varepsilon_{\partial\cl{J}}$ to $50\%$ and the stopping criteria $\Delta \varepsilon_{\partial\cl{I}}$ and $\Delta \varepsilon_{\partial\cl{J}}$ to $10^{-3}$. For the fractional Sobolev regularization, we selected $\epsilon_{\mathrm{d}}=\epsilon_{\mathrm{c}}=10^{-4}$ and $k_{\mathrm{d}}=k_{\mathrm{c}}=1.5$. After $15$ outer iterations, we obtained the index sets in Figure \ref{fg:index.ex5} with $149$ (Figure \ref{fg:ex5.div.index}) and $103$ (Figure \ref{fg:ex5.curl.index}) entries for the divergence- and curl-free fields, respectively.

\begin{figure}
\centering
\includegraphics[clip,trim={0 0 0 0},width=0.65\textwidth]{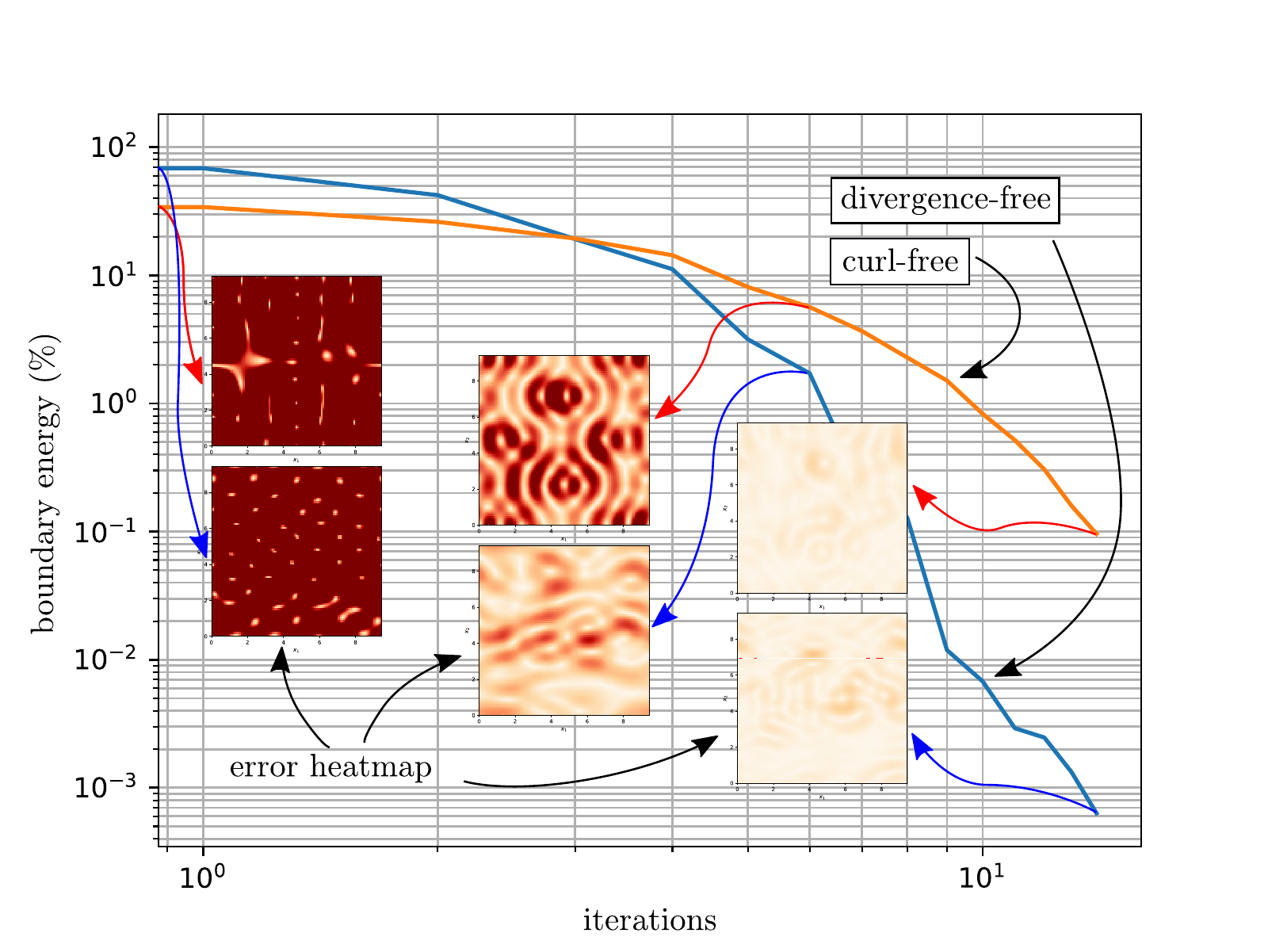}
\caption{Convergence of the relative energy boundary along with error heatmaps for iterations 1, 8, and 14 for both divergence- and curl-free errors. Blue curve: divergence-free. Orange curve: curl-free.}
\label{fg:ex5.conv}
\end{figure}

\begin{figure}
\centering
\subfloat[Underlying field]{\label{fg:ex5.total.anal}\includegraphics[clip,trim={60 0 60 0},width=0.33\textwidth]{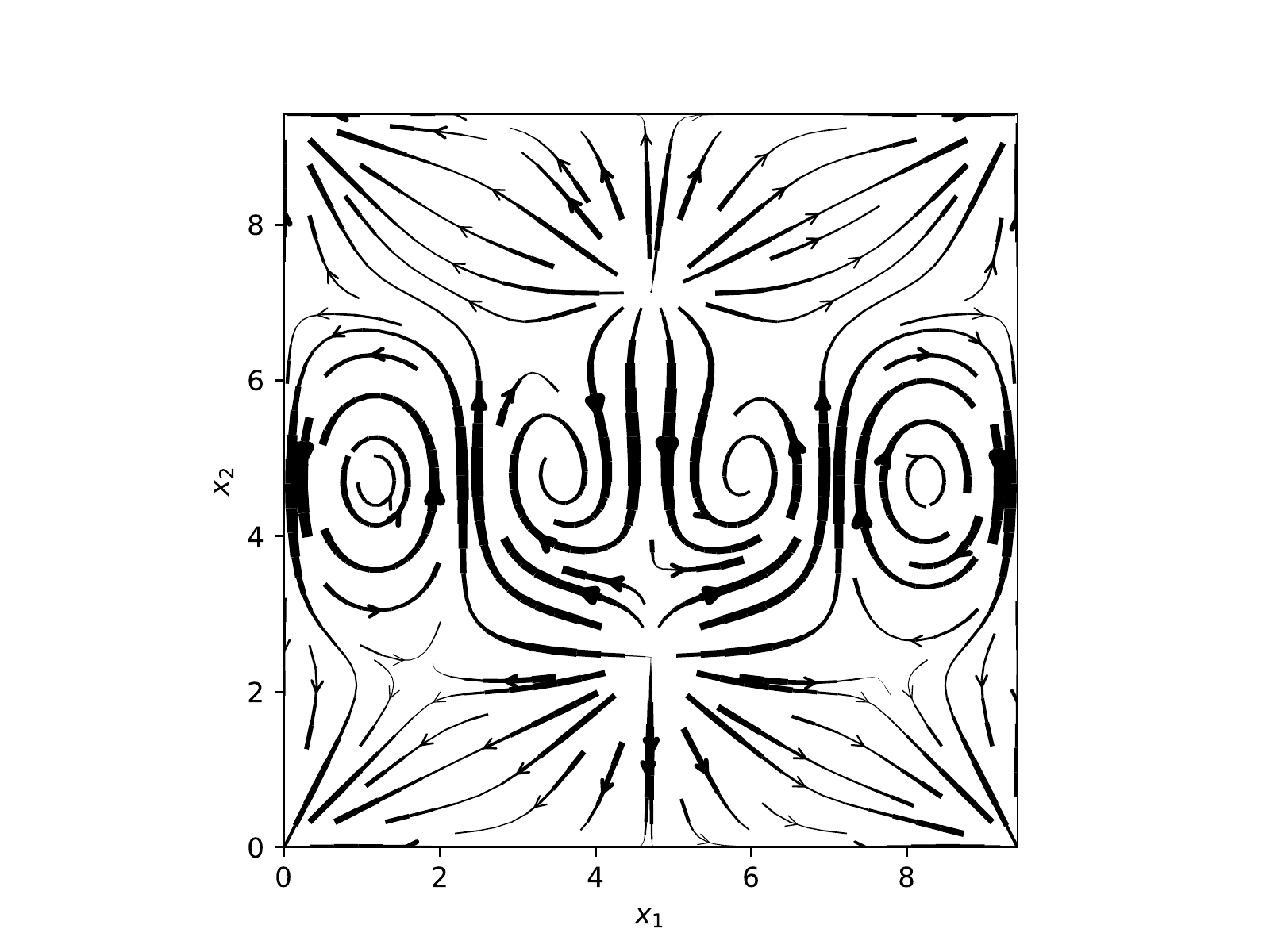}}
\subfloat[Underlying div-free field]{\label{fg:ex5.div.anal}\includegraphics[clip,trim={60 0 60 0},width=0.33\textwidth]{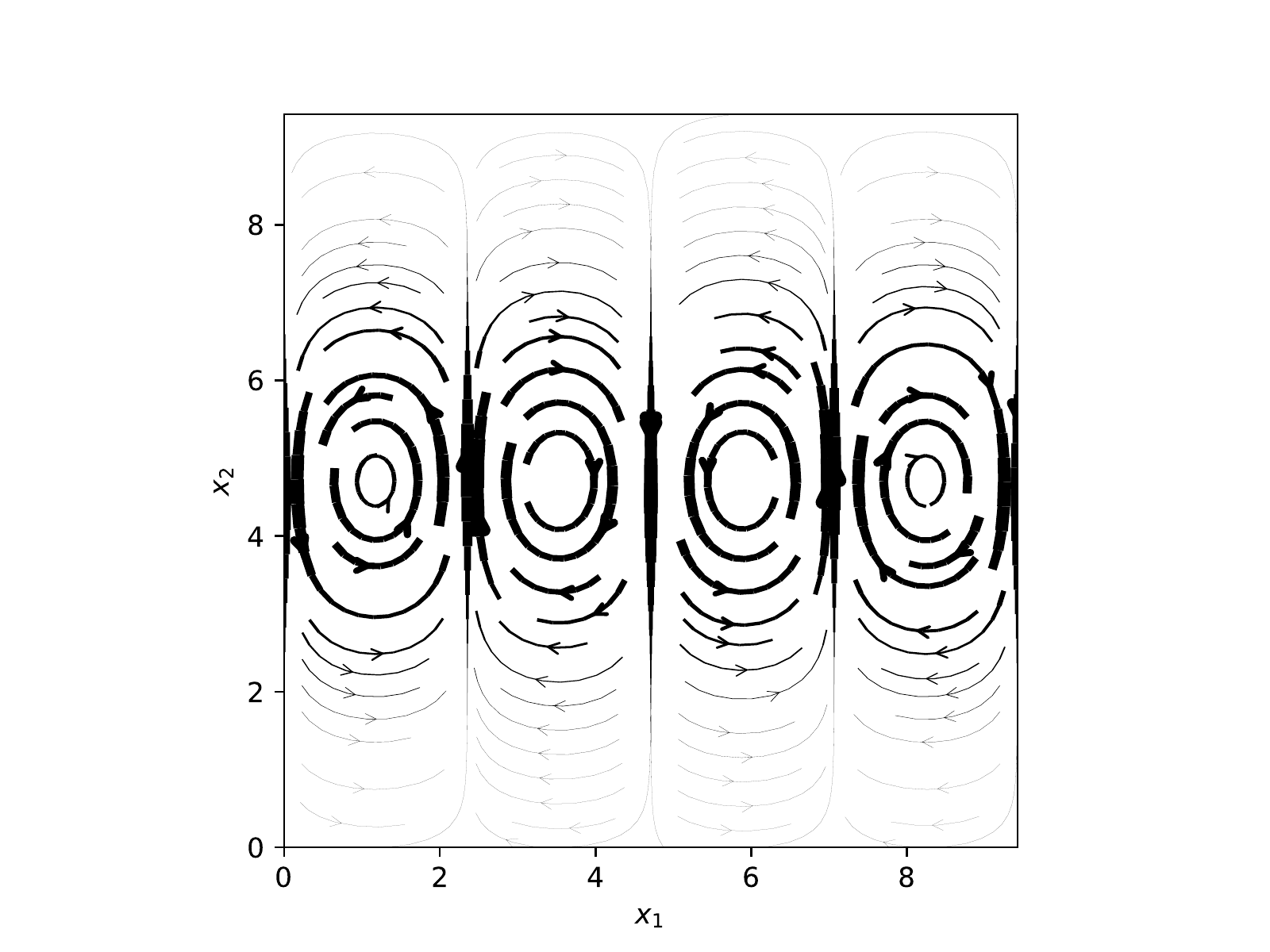}}
\subfloat[Underlying curl-free field]{\label{fg:ex5.curl.anal}\includegraphics[clip,trim={60 0 60 0},width=0.33\textwidth]{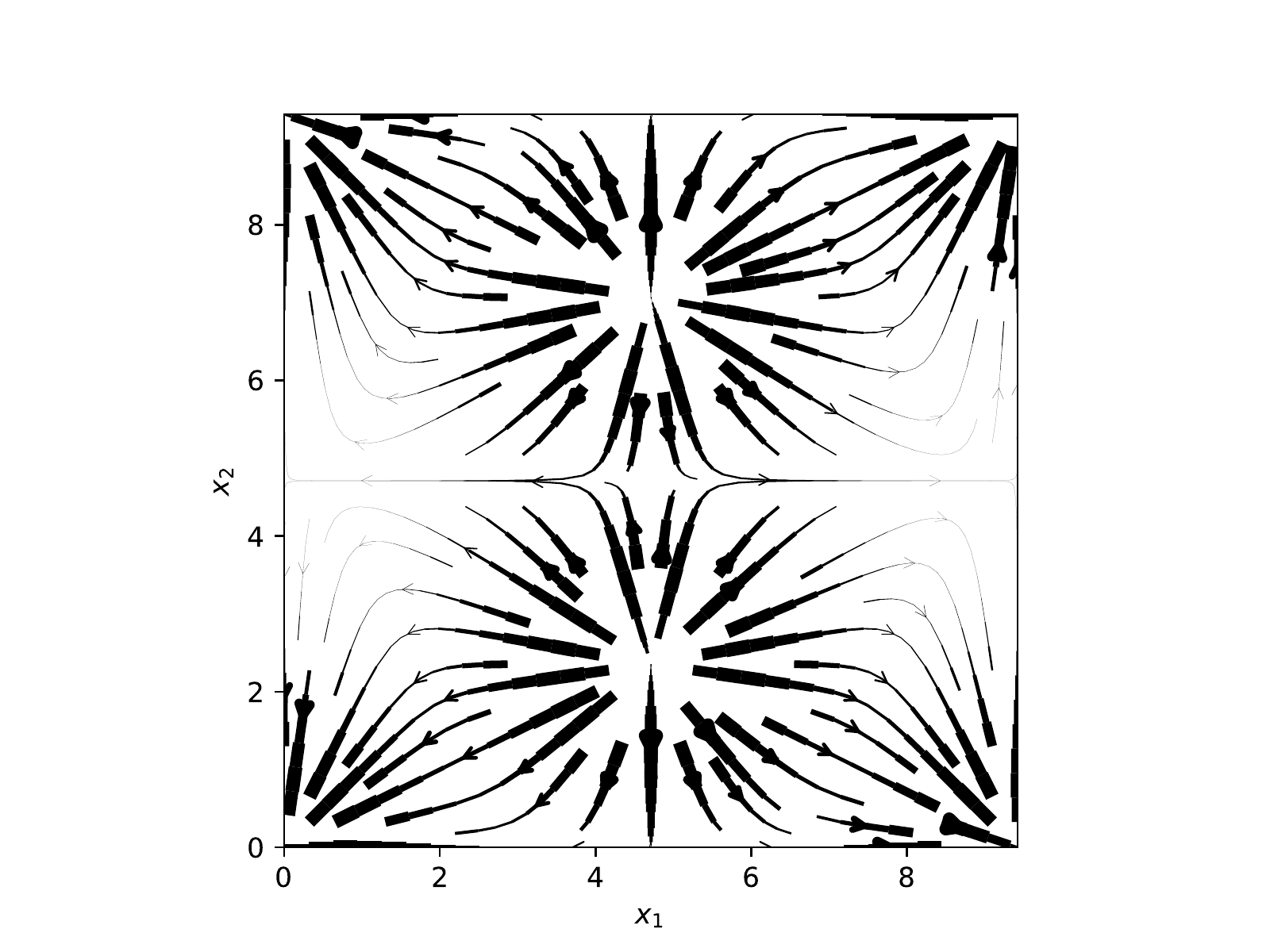}} \\
\subfloat[\PiSL\, field reconstruction]{\label{fg:ex5.total.PiSL}\includegraphics[clip,trim={60 0 60 0},width=0.33\textwidth]{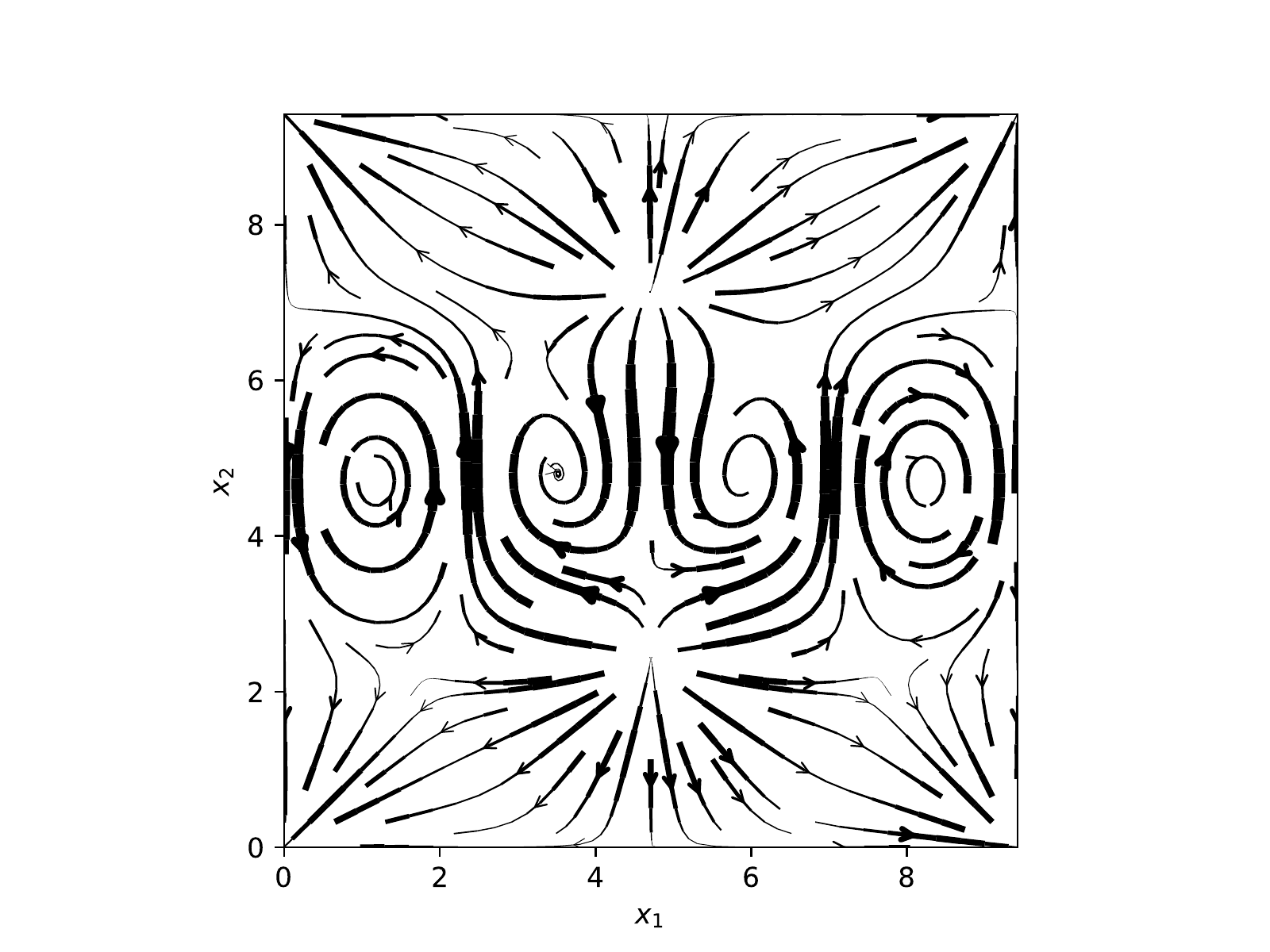}}
\subfloat[\PiSL\, div-free reconstruction]{\label{fg:ex5.div.PiSL}\includegraphics[clip,trim={60 0 60 0},width=0.33\textwidth]{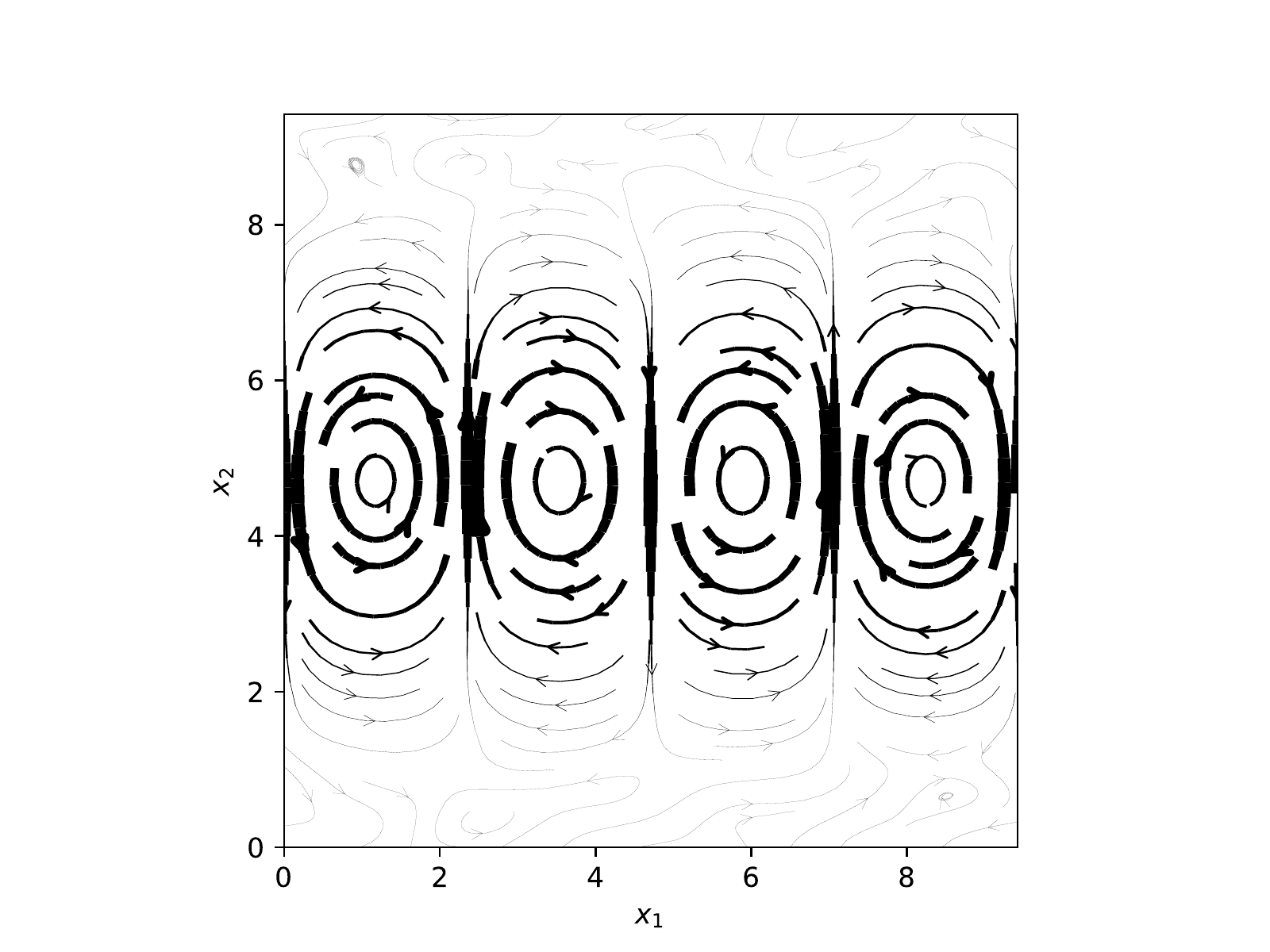}}
\subfloat[\PiSL\, curl-free reconstruction]{\label{fg:ex5.curl.PiSL}\includegraphics[clip,trim={60 0 60 0},width=0.33\textwidth]{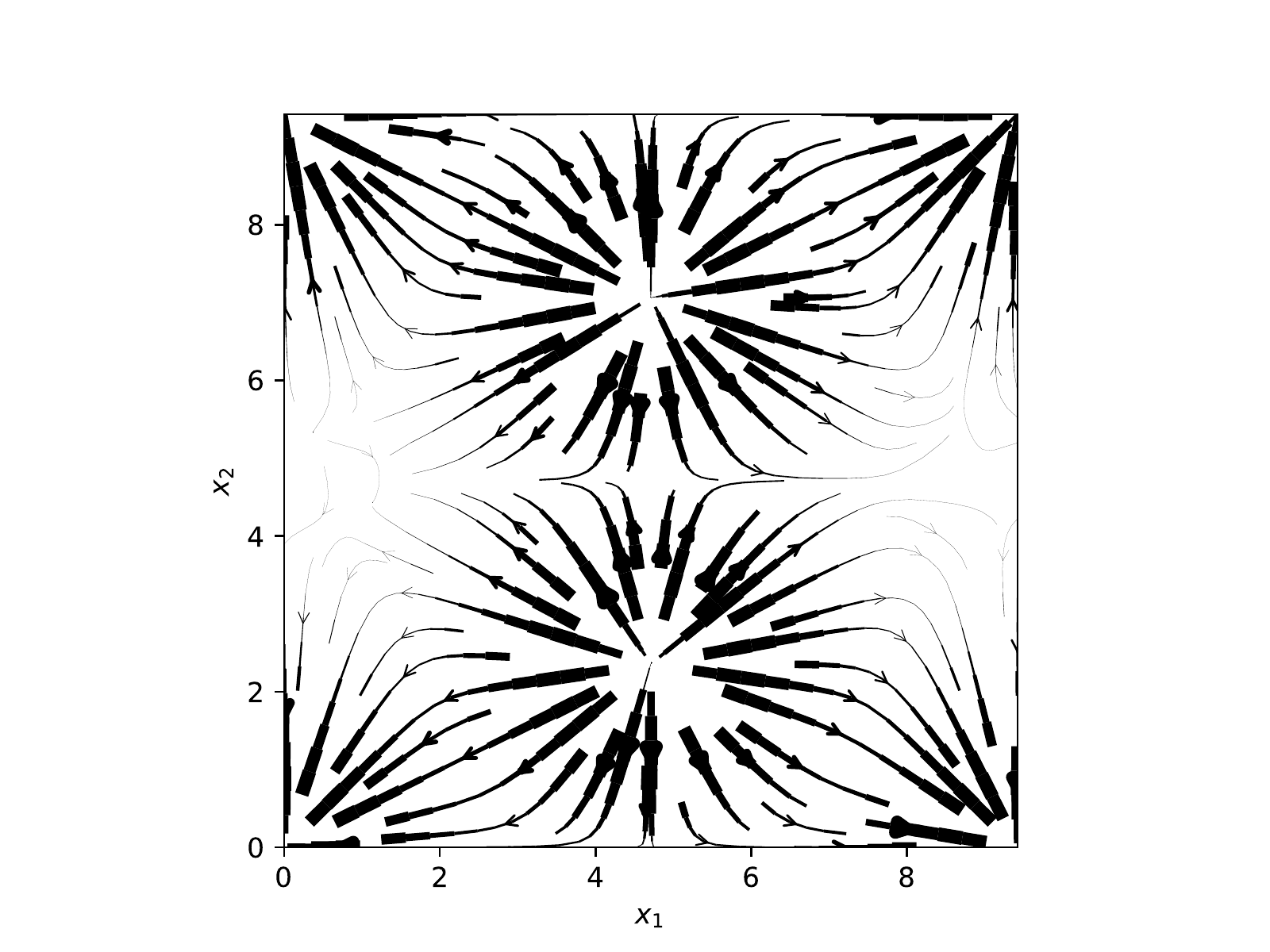}} \\
\subfloat[Total error reconstruction]{\label{fg:ex5.total.err}\includegraphics[clip,trim={0 0 0 0},width=0.33\textwidth]{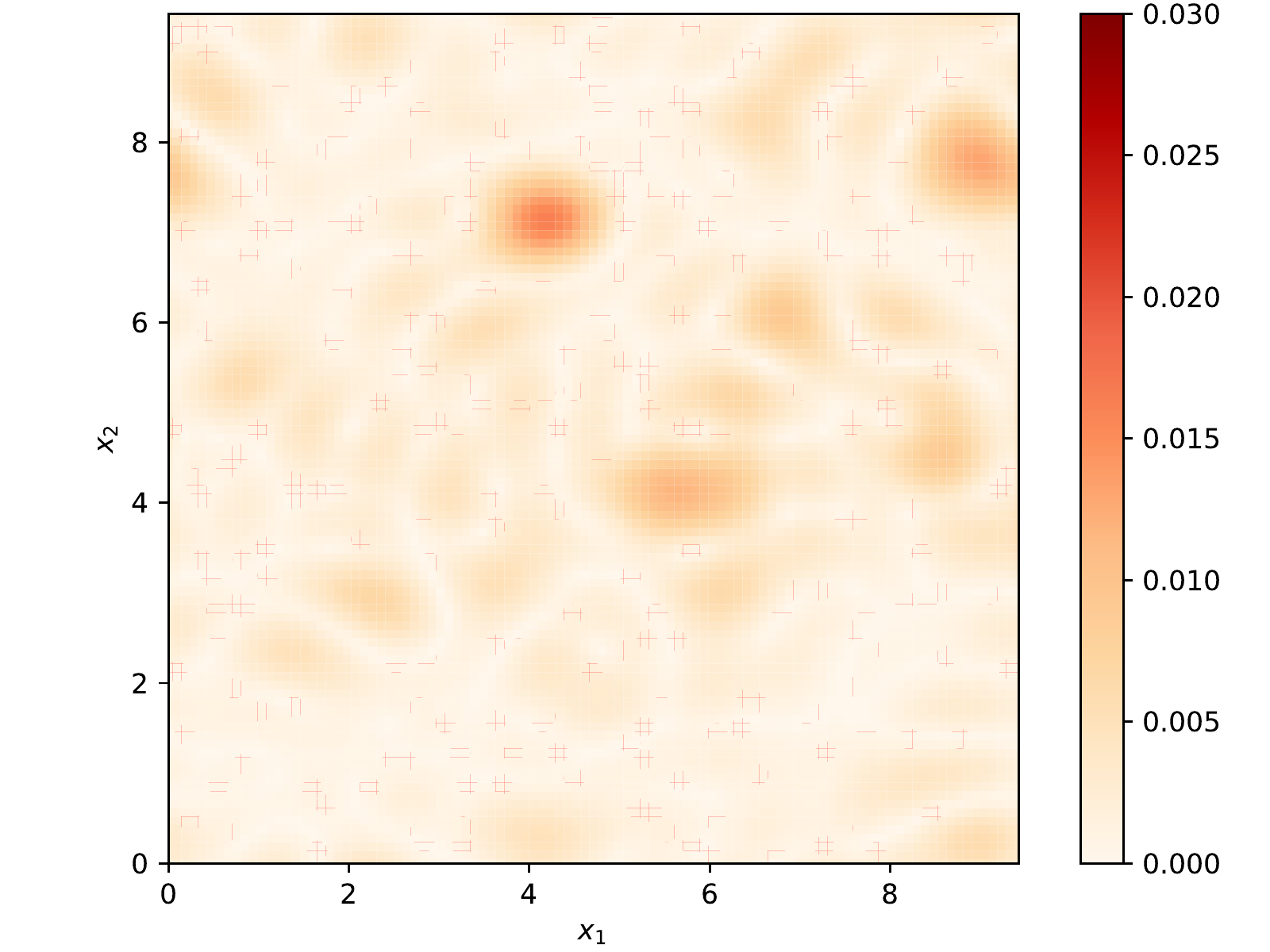}}
\subfloat[Div-free error]{\label{fg:ex5.div.err}\includegraphics[clip,trim={0 0 0 0},width=0.33\textwidth]{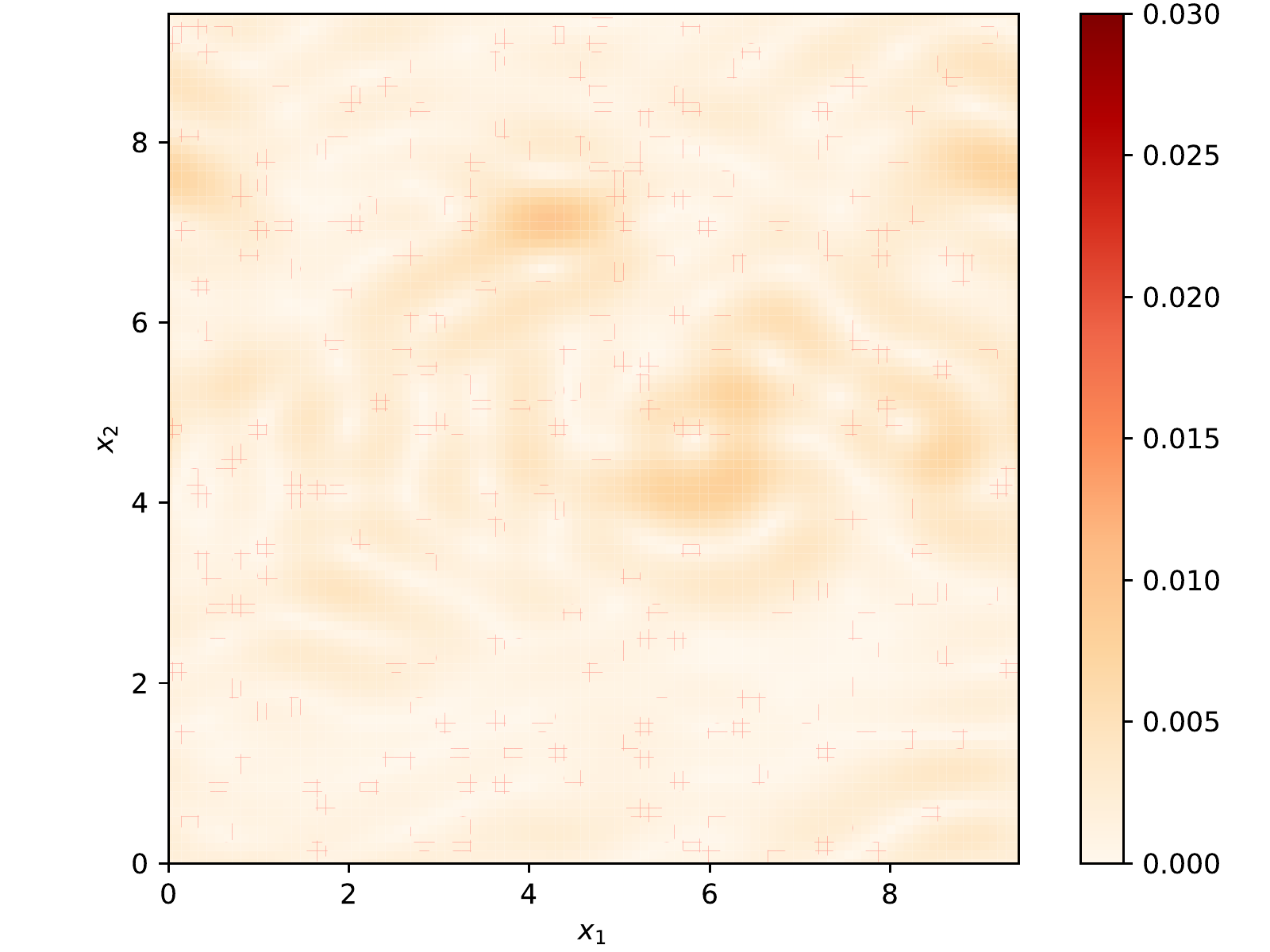}}
\subfloat[Curl-free error]{\label{fg:ex5.curl.err}\includegraphics[clip,trim={0 0 0 0},width=0.33\textwidth]{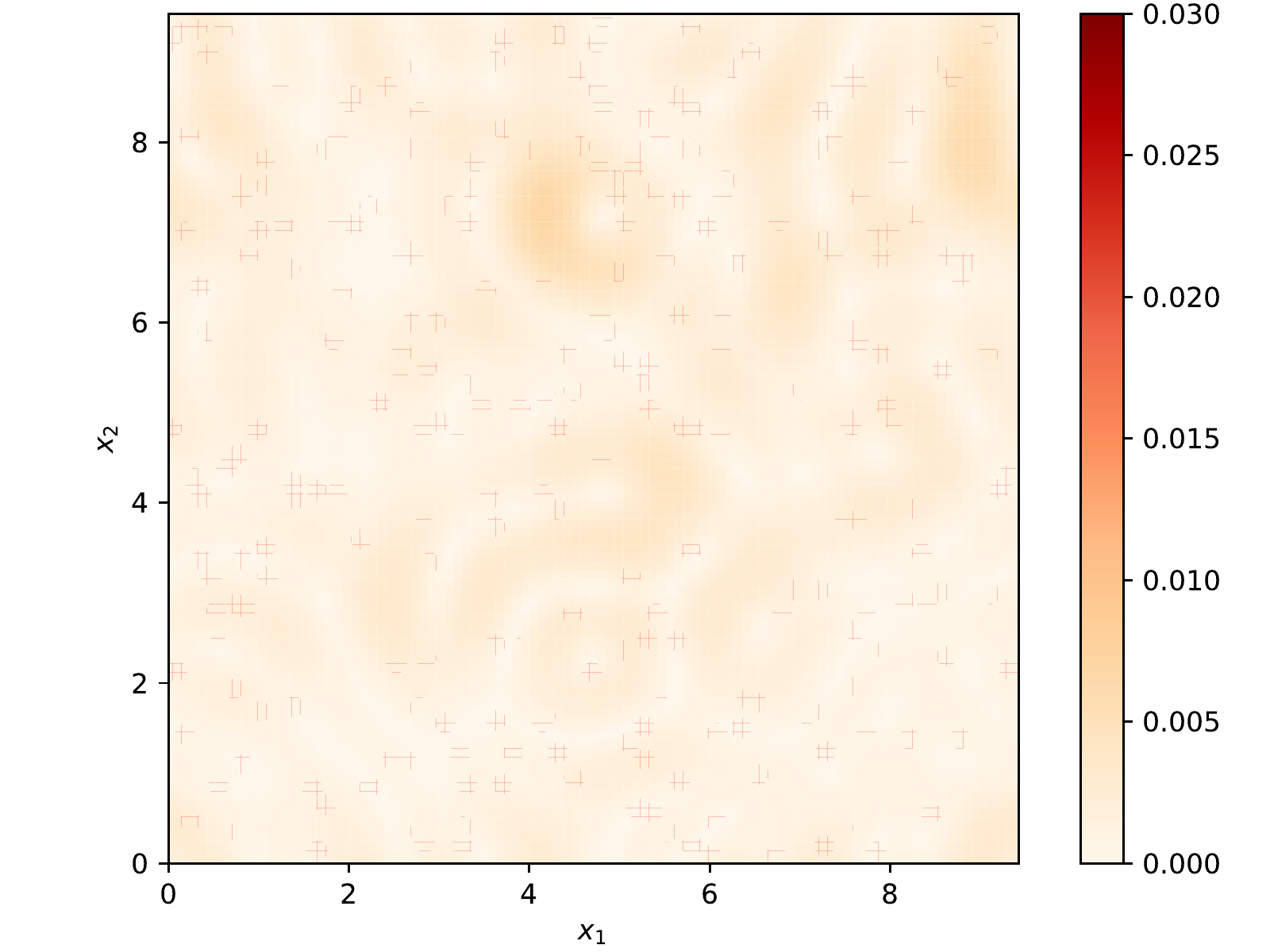}}
\caption{Reconstructed \PiSL\, curl- and divergence-free fields. Panels \protect\subref{fg:ex5.total.anal}: $\bs{u}$ field, \protect\subref{fg:ex5.div.anal}: $\bs{u}_{\mathrm{d}}$ field, and \protect\subref{fg:ex5.curl.anal}: $\bs{u}_{\mathrm{c}}$ field. Panels \protect\subref{fg:ex5.total.PiSL}: $\bsuI+\bssJ$ field, \protect\subref{fg:ex5.div.PiSL}: $\bsuI$ field, and \protect\subref{fg:ex5.curl.PiSL}: $\bssJ$ field. Panels \protect\subref{fg:ex5.total.err}: error in the total field, \protect\subref{fg:ex5.div.err}: error in the divergence-free field, and \protect\subref{fg:ex5.curl.err}: error in the curl-free field.}
\label{fg:fields.ex5}
\end{figure}

Table \ref{tb:ex5} presents the $L^\infty$ and $L^2$ norms of the error $e(\bs{x})$ for the total field $\bsuI^{\mathrm{opt}}+\bssJ^{\mathrm{opt}}$, divergence-free field $\bsuI$, and curl-free field $\bssJ$. As $L^\infty(\|\bs{u}\|) \approx 1.16$, the maximum relative error in any field is less than $2.0\%$. This error is because we assume limited access to the data and can sample the data at a few fixed location points. This bias error vanishes if we allow the data to be sampled randomly at each optimization iteration. Figure \ref{fg:ex5.conv} presents the convergence of the relative energy boundary along with error heatmaps for iterations $1$, $8$, and $14$ for both divergence- and curl-free errors. The blue curve depicts the convergence of the divergence-free reconstruction field, whereas the orange curve represents the convergence of the curl-free reconstruction field. The index sets increase with the iterations, these curves suggest that \PiSL\ is endowed with spectral convergence. That is, for a constant $q$, we obtain $|\bs{\upsilon}_{\bs{\alpha}}|$ and $|\bs{\varsigma}_{\bs{\alpha}}| \sim \cl{O} (\mathrm{e}^{-|\bs{\alpha}|^b})$ where $b$ is the exponential index of convergence. For this problem, the exponential index of convergence is $b \approx -20$. In Figure \ref{fg:fields.ex5}, from left to right in the first row, we present the underlying analytical fields: $\bs{u}$, $\bs{u}_{\mathrm{c}}$, and $\bs{u}_{\mathrm{d}}$. From left to right in the second row, we display the reconstructions: $\bsuI+\bssJ$, $\bsuI$, and $\bssJ$. From left to right in the third row, the reconstruction error corresponding to $\bsuI+\bssJ$, $\bsuI$, and $\bssJ$ is in a scale from $0$ to $0.03$.

\begin{table}[!h]
\caption{$L^\infty$ and $L^2$ norms of the error fields $e$ for the reconstructed total, curl-, and divergence-free fields.}
\begin{tabular}{ c c c }
 $e(\bs{x})$ & $L^2(e(\bs{x}))$ & $L^\infty(e(\bs{x}))$ \\\hline
 $\norm{\bs{u}(\bs{x})-\bsuI^{\mathrm{opt}}(\bs{x})-\bssJ^{\mathrm{opt}}(\bs{x})}{}$ & $0.3884$ & $0.02349$ \\
 $\norm{\bs{u}_{\mathrm{d}}(\bs{x})-\bsuI^{\mathrm{opt}}(\bs{x})}{}$ & $0.3354$ & $0.02008$ \\
 $\norm{\bs{u}_{\mathrm{c}}(\bs{x})-\bssJ^{\mathrm{opt}}(\bs{x})}{}$ & $0.2376$ & $0.01631$ \\
\end{tabular}
\label{tb:ex5}
\end{table}

\FloatBarrier

\subsection{Two-dimensional \PiSL: vortices with a different scale and source}

In the domain $\cl{D} = [D,D]$ with $D = 2\pi$, consider the underlying divergence-free field
\begin{equation}
\bs{u}_{\mathrm{d}}(\bs{x}) = \fr{1}{2} (\cos(x_1)\sin(x_2) + \cos(2x_1)\sin(2x_2),-\sin(x_1)\cos(x_2)-\sin(2x_1)\cos(2x_2)).
\end{equation}
In addition, based on \eqref{eq:scalar.function} and \eqref{eq:analytical.field.construction}$_2$, consider the underlying curl-free field
\begin{equation}
\bs{u}_{\mathrm{c}}(\bs{x}) = - \bs{\upsilon}_{\mathrm{c}} (\bs{x};\bs{x}_0),
\end{equation}
where $\bs{x}_0 = (D/2,D/2)$. Then, the field we aim to reconstruct while performing the discrete $L^2$ HHD is given by
\begin{equation}
\bs{u}(\bs{x}) = \bs{u}_{\mathrm{d}}(\bs{x}) + \bs{u}_{\mathrm{c}}(\bs{x}).
\end{equation}

\begin{figure}[!h]
\centering
\subfloat[Index set: divergence-free field]{\label{fg:ex6.div.index}\qquad\qquad\qquad\includegraphics[]{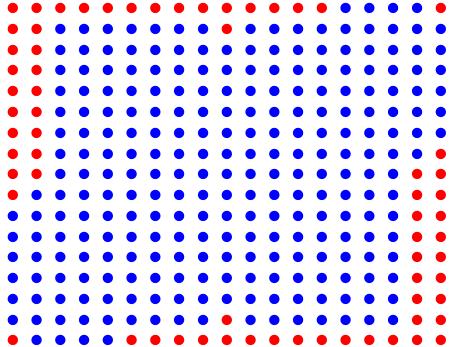}\qquad\qquad\qquad} \hspace{-2cm}
\subfloat[Index set: curl-free field]{\label{fg:ex6.curl.index}\qquad\qquad\qquad\includegraphics{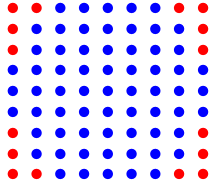}\qquad\qquad\qquad}
\caption{Panels \protect\subref{fg:ex6.div.index} and \protect\subref{fg:ex6.curl.index}: Index sets of the reconstructed \PiSL\, divergence- and curl-free fields, respectively. Blue dots represent the indices retained in the Fourier construction. Red dots represent indices removed from the Fourier construction.}
\label{fg:index.ex6}
\end{figure}

The discrete $L^2$ HHD is carried out using $250$ fixed measurements at random points in the domain $\cl{D}=[0,2\pi]^2$. We set the residual boundary energy $\varepsilon_{\partial\cl{I}}$ and $\varepsilon_{\partial\cl{J}}$ to $50\%$ and the stopping criteria $\Delta \varepsilon_{\partial\cl{I}}$ and $\Delta \varepsilon_{\partial\cl{J}}$ to $10^{-3}$. For the fractional Sobolev regularization, we selected $\epsilon_{\mathrm{d}}=\epsilon_{\mathrm{c}}=10^{-4}$ and $k_{\mathrm{d}}=k_{\mathrm{c}}=1.5$. After $11$ outer iterations, we obtained the index sets in Figure \ref{fg:index.ex6} with $257$ (Figure \ref{fg:ex6.div.index}) and $65$ (Figure \ref{fg:ex6.curl.index}) entries for the divergence- and curl-free fields, respectively.

\begin{figure}
\centering
\subfloat[Underlying field]{\label{fg:ex6.total.anal}\includegraphics[clip,trim={60 0 60 0},width=0.33\textwidth]{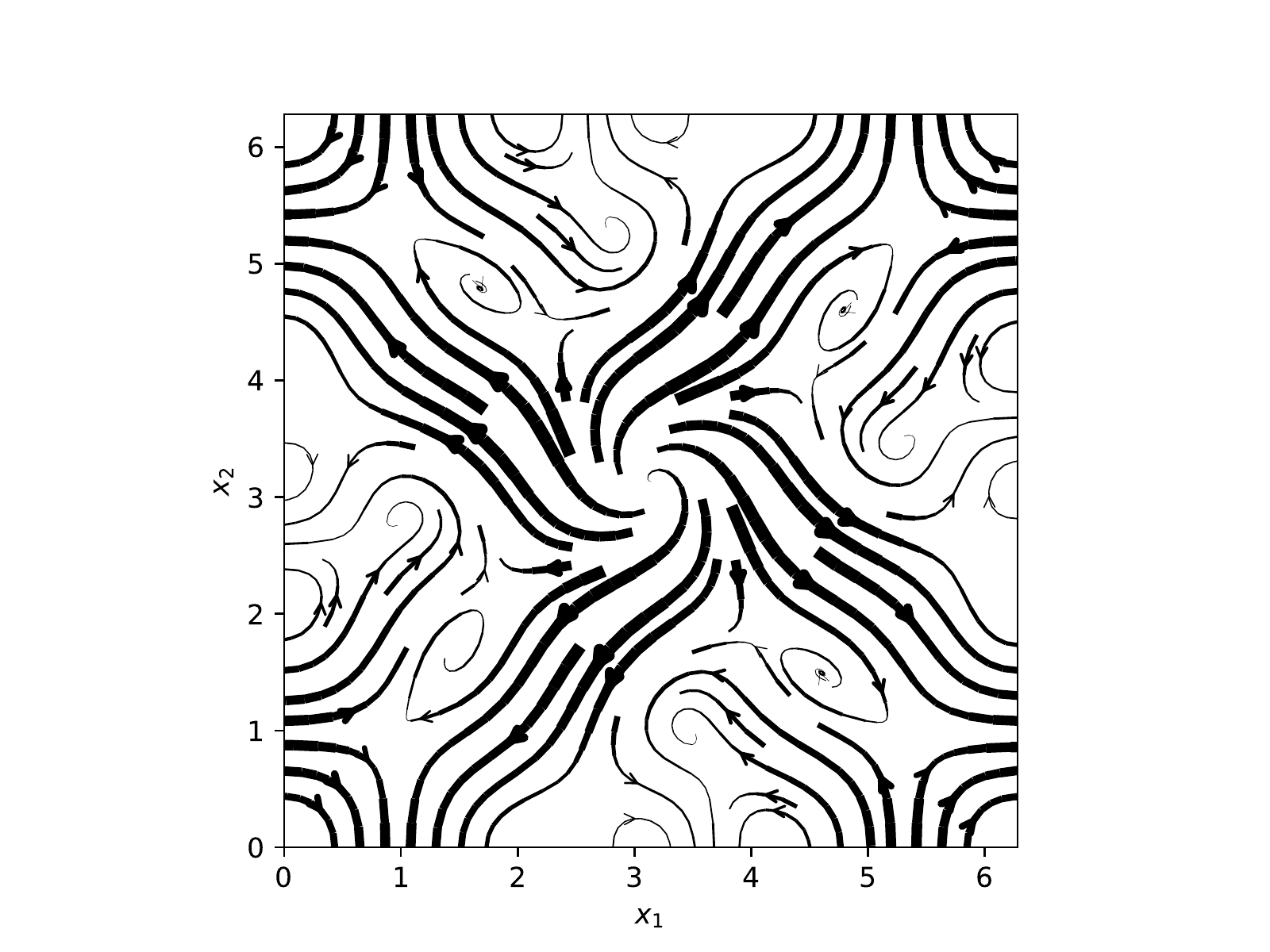}}
\subfloat[Underlying div-free field]{\label{fg:ex6.div.anal}\includegraphics[clip,trim={60 0 60 0},width=0.33\textwidth]{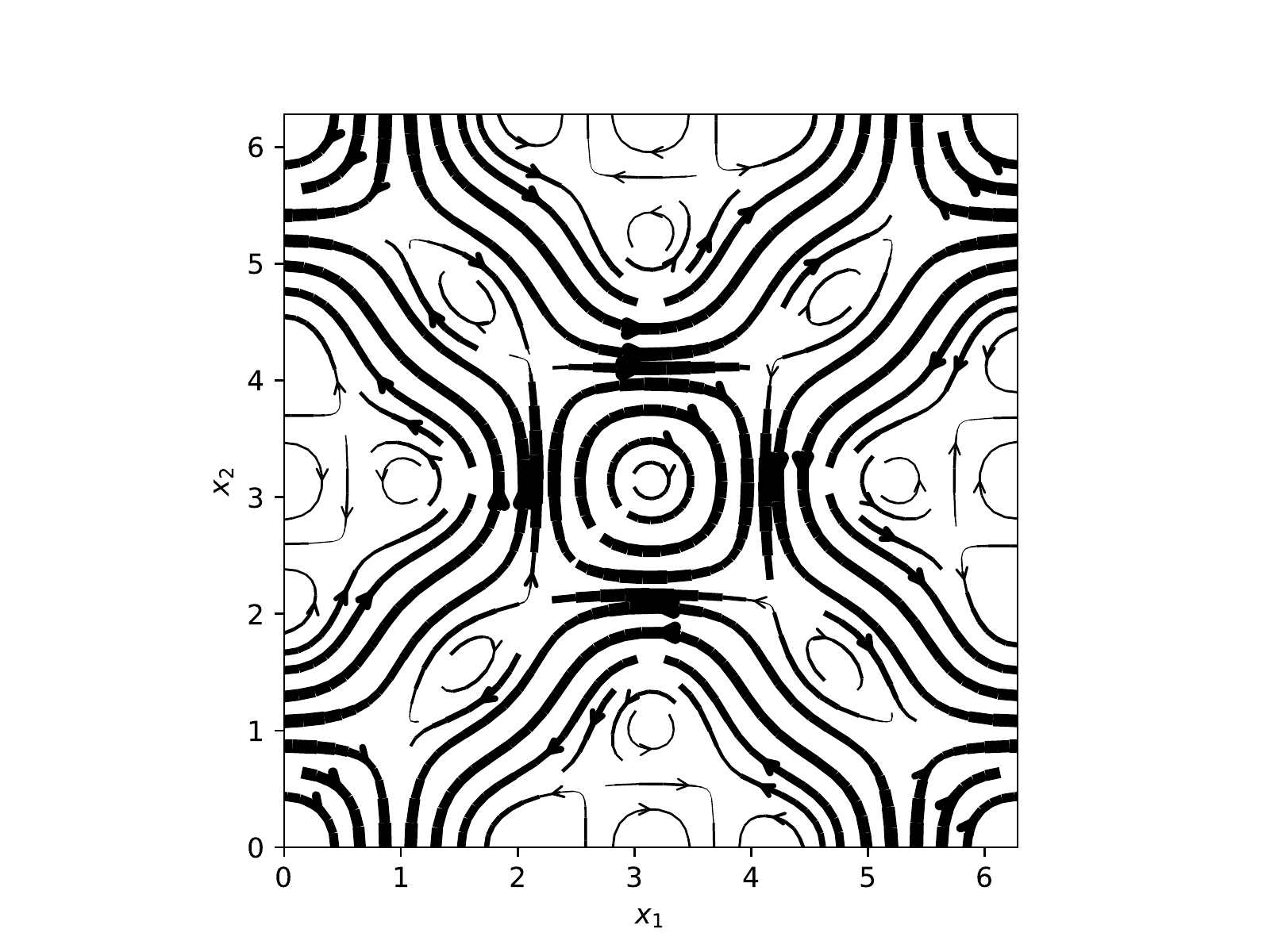}}
\subfloat[Underlying curl-free field]{\label{fg:ex6.curl.anal}\includegraphics[clip,trim={60 0 60 0},width=0.33\textwidth]{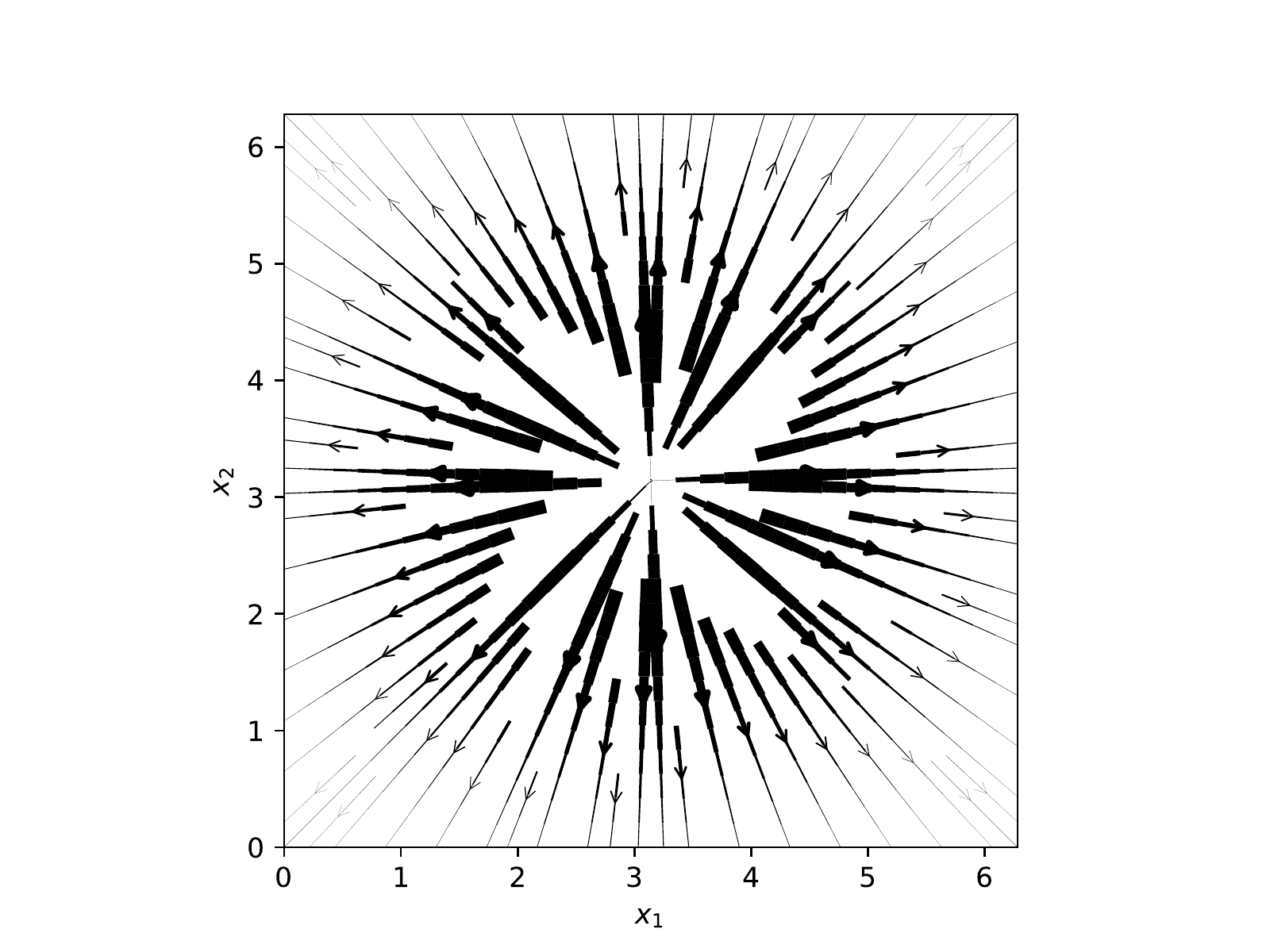}} \\
\subfloat[\PiSL\, field reconstruction]{\label{fg:ex6.total.PiSL}\includegraphics[clip,trim={60 0 60 0},width=0.33\textwidth]{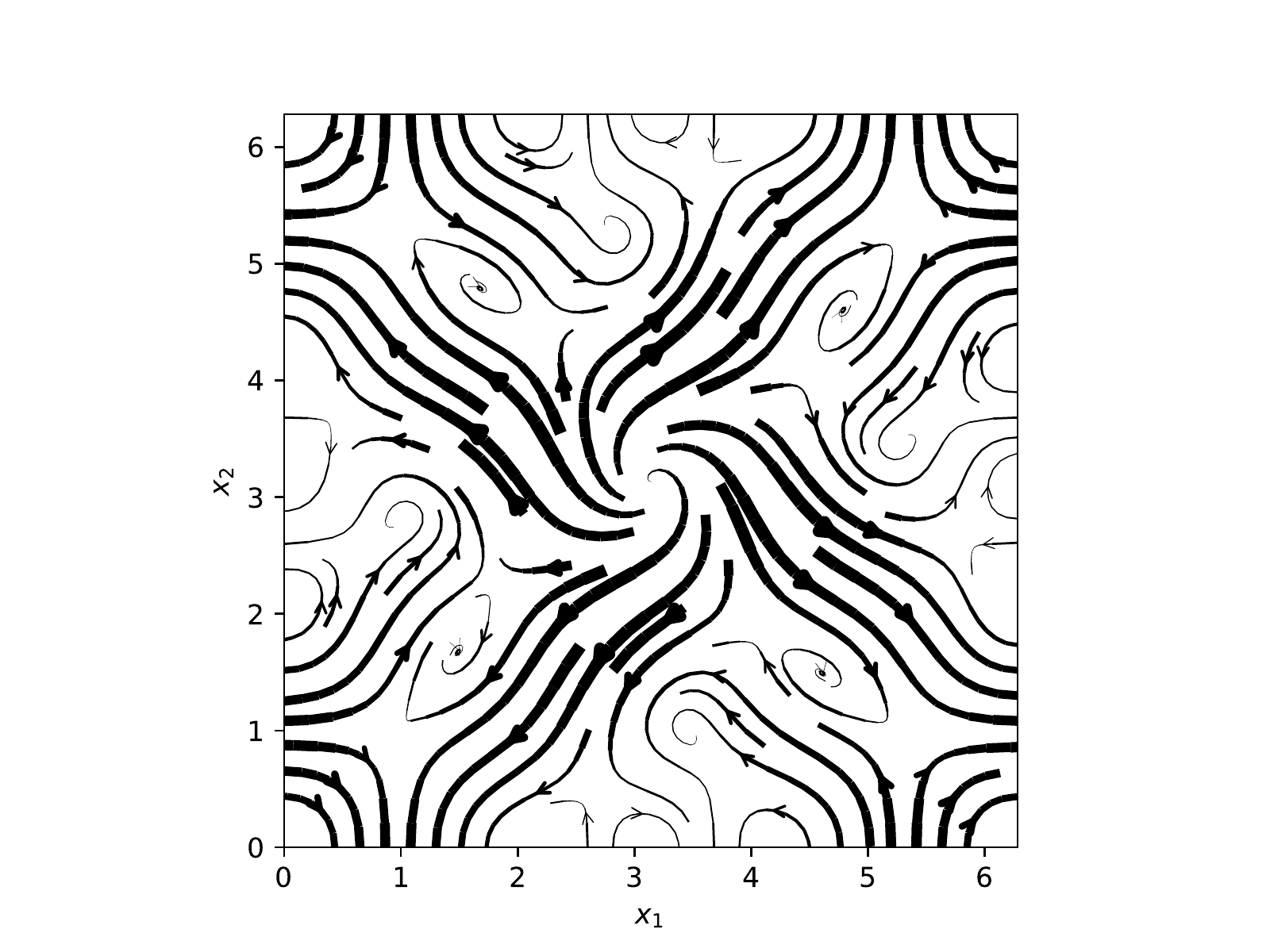}}
\subfloat[\PiSL\, div-free reconstruction]{\label{fg:ex6.div.PiSL}\includegraphics[clip,trim={60 0 60 0},width=0.33\textwidth]{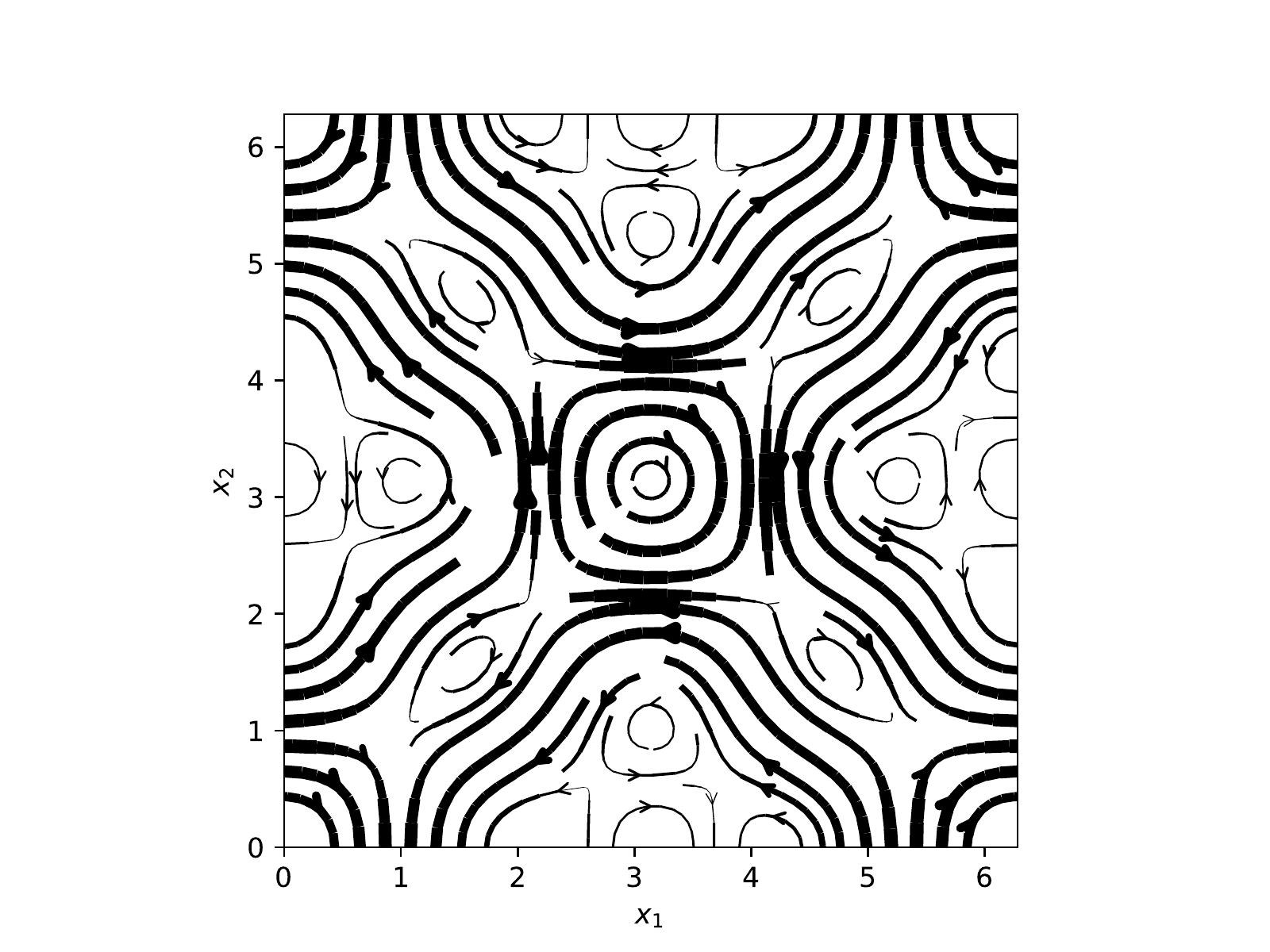}}
\subfloat[\PiSL\, curl-free reconstruction]{\label{fg:ex6.curl.PiSL}\includegraphics[clip,trim={60 0 60 0},width=0.33\textwidth]{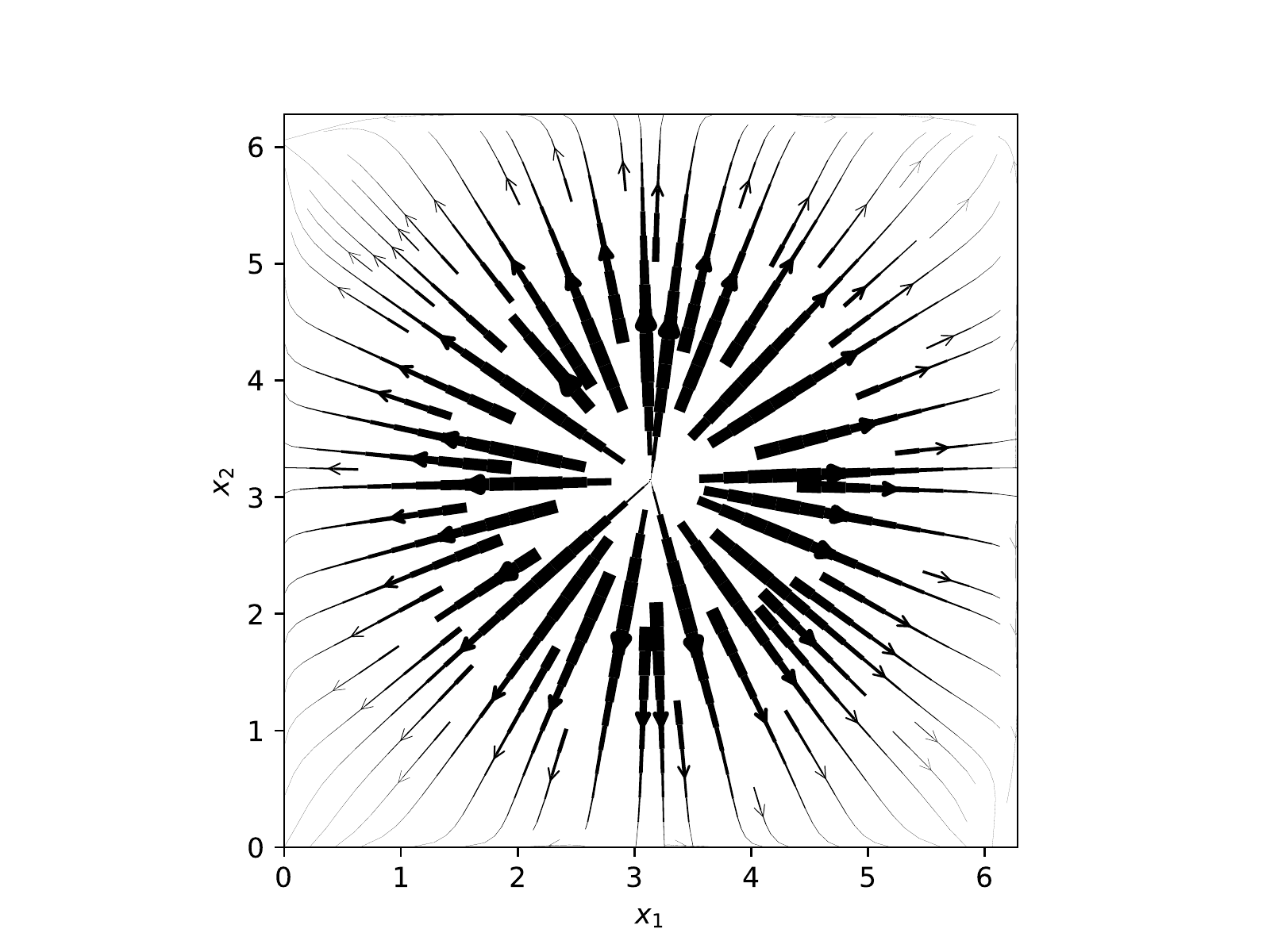}} \\
\subfloat[Total error reconstruction]{\label{fg:ex6.total.err}\includegraphics[clip,trim={0 0 0 0},width=0.33\textwidth]{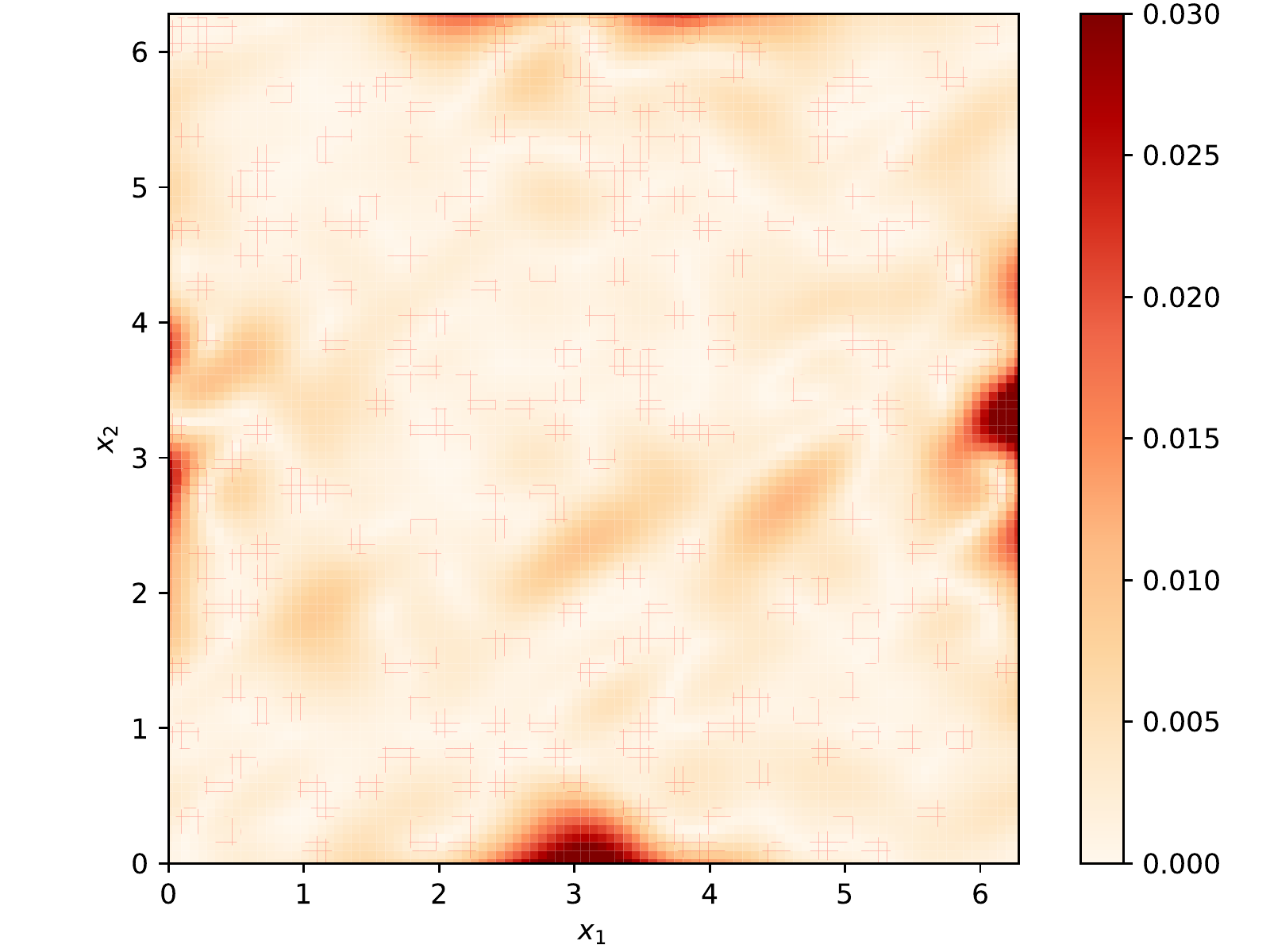}}
\subfloat[Div-free error]{\label{fg:ex6.div.err}\includegraphics[clip,trim={0 0 0 0},width=0.33\textwidth]{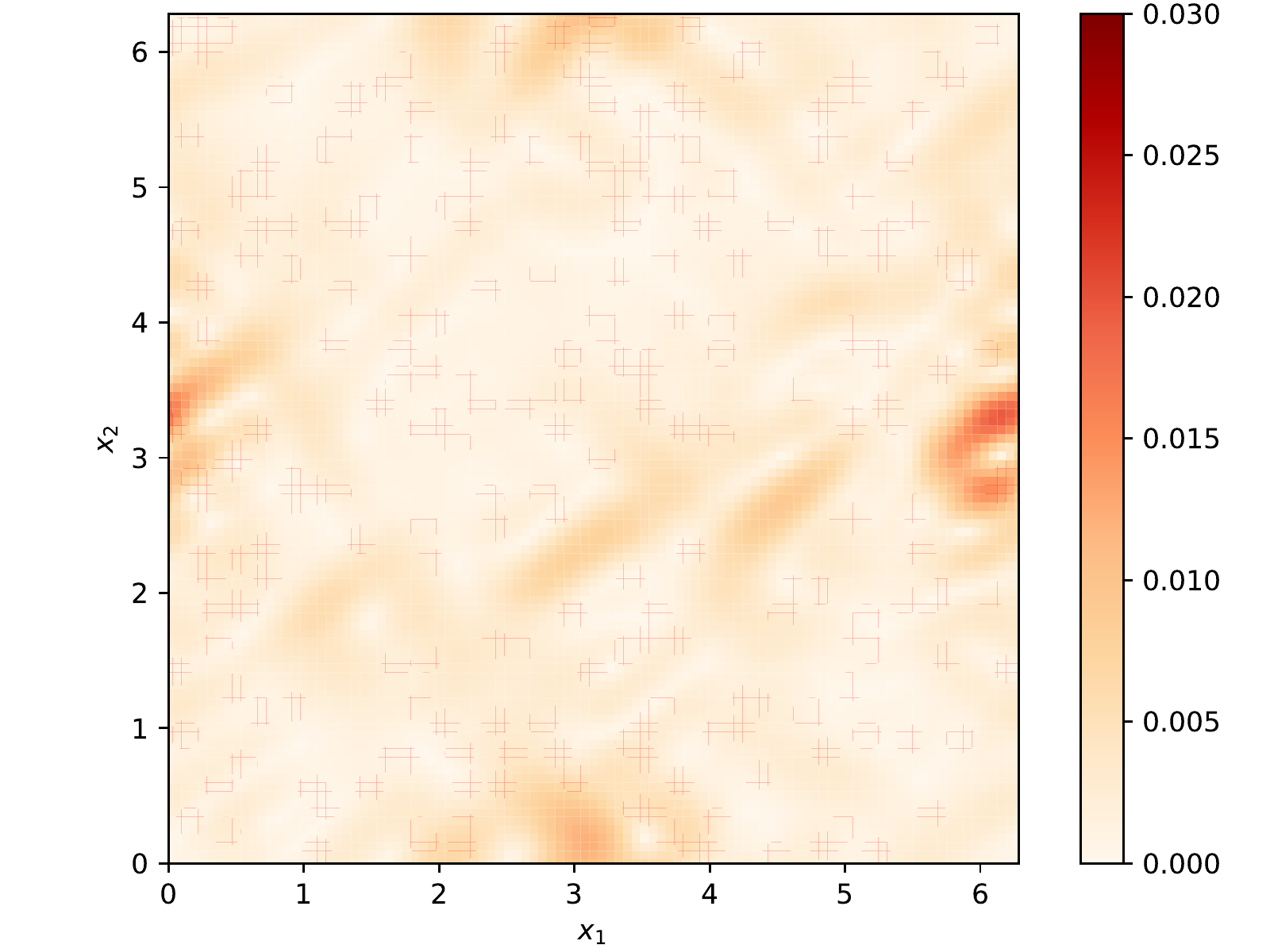}}
\subfloat[Curl-free error]{\label{fg:ex6.curl.err}\includegraphics[clip,trim={0 0 0 0},width=0.33\textwidth]{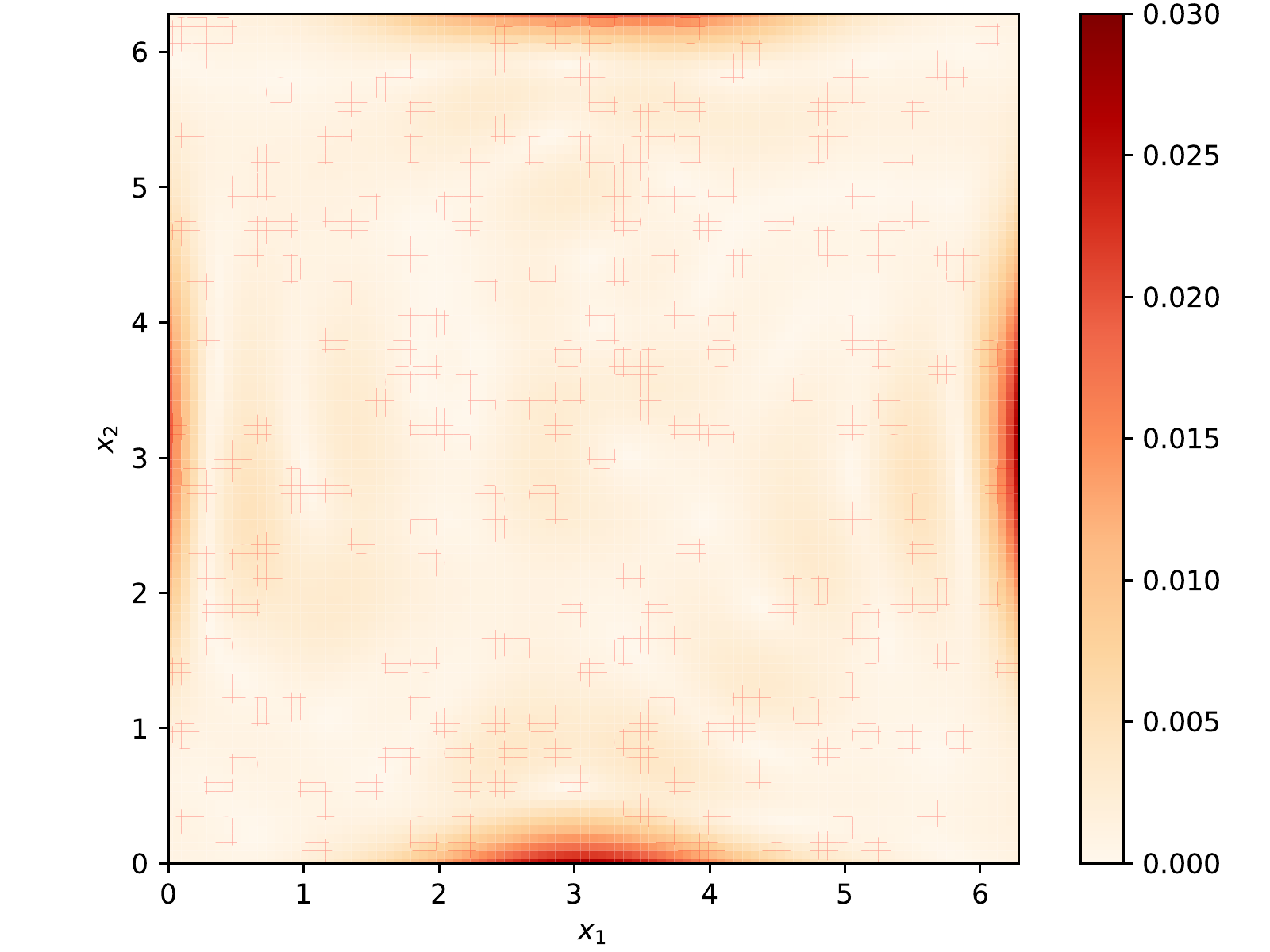}}
\caption{Reconstructed \PiSL\, curl- and divergence-free fields. Panels \protect\subref{fg:ex6.total.anal}: $\bs{u}$ field, \protect\subref{fg:ex6.div.anal}: $\bs{u}_{\mathrm{d}}$ field, and \protect\subref{fg:ex6.curl.anal}: $\bs{u}_{\mathrm{c}}$ field. Panels \protect\subref{fg:ex6.total.PiSL}: $\bsuI+\bssJ$ field, \protect\subref{fg:ex6.div.PiSL}: $\bsuI$ field, and \protect\subref{fg:ex6.curl.PiSL}: $\bssJ$ field. Panels \protect\subref{fg:ex6.total.err}: error in the total field, \protect\subref{fg:ex6.div.err}: error in the divergence-free field, and \protect\subref{fg:ex6.curl.err}: error in the curl-free field.}
\label{fg:fields.ex6}
\end{figure}

Table \ref{tb:ex6} lists the $L^\infty$ and $L^2$ norms of the error $e(\bs{x})$ for the total field $\bsuI^{\mathrm{opt}}+\bssJ^{\mathrm{opt}}$, divergence-free field $\bsuI$, and curl-free field $\bssJ$. As $L^\infty(\|\bs{u}\|) \approx 0.97$, the maximum relative error in any field is less than $2.5\%$. In Figure \ref{fg:fields.ex6}, from left to right in the first row, we present the underlying analytical fields: $\bs{u}$, $\bs{u}_{\mathrm{c}}$, and $\bs{u}_{\mathrm{d}}$. From left to right in the second row, we display the reconstructions: $\bsuI+\bssJ$, $\bsuI$, and $\bssJ$. From left to right in the third row, the reconstruction error corresponding to $\bsuI+\bssJ$, $\bsuI$, and $\bssJ$ is in a scale from $0$ to $0.03$.
\begin{table}[!h]
\caption{$L^\infty$ and $L^2$ norms of the error fields $e$ for the reconstructed total, curl-, and divergence-free fields.}
\begin{tabular}{ c c c }
 $e(\bs{x})$ & $L^2(e(\bs{x}))$ & $L^\infty(e(\bs{x}))$ \\\hline
 $\norm{\bs{u}(\bs{x})-\bsuI^{\mathrm{opt}}(\bs{x})-\bssJ^{\mathrm{opt}}(\bs{x})}{}$ & $0.9943$ & $0.02519$ \\
 $\norm{\bs{u}_{\mathrm{d}}(\bs{x})-\bsuI^{\mathrm{opt}}(\bs{x})}{}$ & $0.4018$ & $0.01765$ \\
 $\norm{\bs{u}_{\mathrm{c}}(\bs{x})-\bssJ^{\mathrm{opt}}(\bs{x})}{}$ & $0.9566$ & $0.02010$ \\
\end{tabular}
\label{tb:ex6}
\end{table}

\FloatBarrier

\subsection{Three-dimensional \PiSL: vortices with a different scale and source}

This example is a three-dimensional extension of the previous example with a domain $\cl{D} = [D,D,D]$ with $D = 2\pi$. The underlying divergence-free field is given by
\begin{align}
\bs{u}_{\mathrm{d}}(\bs{x}) ={}& (\fr{1}{2}(\cos(x_1)\sin(x_2)\cos(x_3) + \cos(2x_1)\sin(2x_2)\cos(x_3)), \nonumber \\[4pt]
&\fr{1}{2}(-\sin(x_1)\cos(x_2)\cos(x_3) - \sin(2x_1)\cos(2x_2)\cos(2x_3)), \nonumber \\[4pt]
&\fr{1}{2}(\sin(x_1)\sin(x_2)(\sin(x_3) - \fr{1}{2}\sin(2x_3)) + 2\sin(2x_1)\sin(2x_2)(\sin(x_3) \nonumber \\[4pt]
&- \fr{1}{2}\sin(2x_3))) - (\fr{1}{2}\sin(x_3) - \fr{1}{4}\sin(2x_3))\sin(x_1)\sin(x_2)]),
\end{align}
whereas, based on \eqref{eq:scalar.function} and \eqref{eq:analytical.field.construction}$_2$, the curl-free field is
\begin{equation}
\bs{u}_{\mathrm{c}}(\bs{x}) = - \bs{\upsilon}_{\mathrm{c}} (\bs{x};\bs{x}_0),
\end{equation}
where $\bs{x}_0 = (D/2,D/2,D/2)$. Thus, the field we aim to reconstruct while performing the discrete $L^2$ HHD is given by
\begin{equation}
\bs{u}(\bs{x}) = \bs{u}_{\mathrm{d}}(\bs{x}) + \bs{u}_{\mathrm{c}}(\bs{x}).
\end{equation}

\begin{figure}
\centering
\subfloat[$\cl{I}(-4,:,:)$]{\label{fg:index.div.1}\qquad\qquad\qquad\includegraphics[clip,trim={0 0 0 0}]{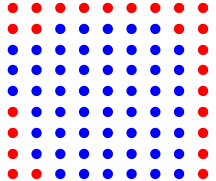}\qquad\qquad\qquad} \hspace{-4cm}
\subfloat[$\cl{I}(-3,:,:)$]{\label{fg:index.div.2}\qquad\qquad\qquad\includegraphics[clip,trim={0 0 0 0}]{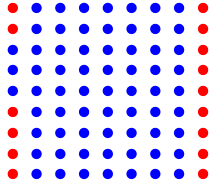}\qquad\qquad\qquad} \hspace{-4cm}
\subfloat[$\cl{I}(-2,:,:)$]{\label{fg:index.div.3}\qquad\qquad\qquad\includegraphics[clip,trim={0 0 0 0}]{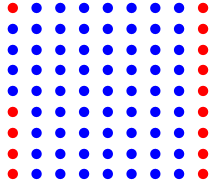}\qquad\qquad\qquad}\\
\subfloat[$\cl{I}(-1,:,:)$]{\label{fg:index.div.4}\qquad\qquad\qquad\includegraphics[clip,trim={0 0 0 0}]{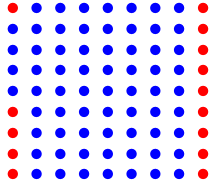}\qquad\qquad\qquad} \hspace{-4cm}
\subfloat[$\cl{I}(0,:,:)$]{\label{fg:index.div.5}\qquad\qquad\qquad\includegraphics[clip,trim={0 0 0 0}]{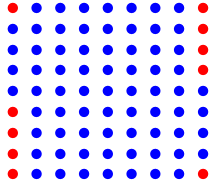}\qquad\qquad\qquad} \hspace{-4cm}
\subfloat[$\cl{I}(1,:,:)$]{\label{fg:index.div.6}\qquad\qquad\qquad\includegraphics[clip,trim={0 0 0 0}]{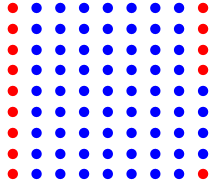}\qquad\qquad\qquad}\\
\subfloat[$\cl{I}(2,:,:)$]{\label{fg:index.div.7}\qquad\qquad\qquad\includegraphics[clip,trim={0 0 0 0}]{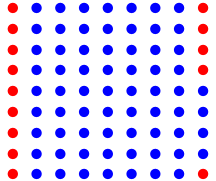}\qquad\qquad\qquad} \hspace{-4cm}
\subfloat[$\cl{I}(3,:,:)$]{\label{fg:index.div.8}\qquad\qquad\qquad\includegraphics[clip,trim={0 0 0 0}]{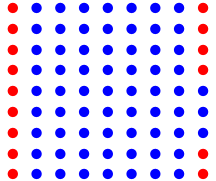}\qquad\qquad\qquad} \hspace{-4cm}
\subfloat[$\cl{I}(4,:,:)$]{\label{fg:index.div.9}\qquad\qquad\qquad\includegraphics[clip,trim={0 0 0 0}]{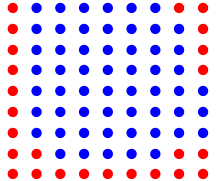}\qquad\qquad\qquad}
\caption{Index set: divergence-free field. Panels \protect\subref{fg:index.div.1}, \protect\subref{fg:index.div.2}, \protect\subref{fg:index.div.3}, \protect\subref{fg:index.div.4}, \protect\subref{fg:index.div.5}, \protect\subref{fg:index.div.6}, \protect\subref{fg:index.div.7}, \protect\subref{fg:index.div.8}, and \protect\subref{fg:index.div.9}: $\cl{I}(-4,:,:)$, $\cl{I}(-3,:,:)$, $\cl{I}(-2,:,:)$, $\cl{I}(-1,:,:)$, $\cl{I}(0,:,:)$, $\cl{I}(1,:,:)$, $\cl{I}(2,:,:)$, $\cl{I}(3,:,:)$, and $\cl{I}(4,:,:)$, respectively. Blue dots represent the indices retained in the Fourier construction. Red dots represent indices removed from the Fourier construction.}
\label{fg:index.div.ex2}
\end{figure}

The discrete $L^2$ HHD is carried out using $1024$ fixed measurements at random points in the domain $\cl{D}=[0,2\pi]^3$. We set the residual boundary energies $\varepsilon_{\partial\cl{I}}$ and $\varepsilon_{\partial\cl{J}}$ to $20\%$ and the stopping criteria $\Delta \varepsilon_{\partial\cl{I}}$ and $\Delta \varepsilon_{\partial\cl{J}}$ to $10^{-3}$. For the fractional Sobolev regularization, we selected $\epsilon_{\mathrm{d}}=\epsilon_{\mathrm{c}}=10^{-6}$ and $k_{\mathrm{d}}=k_{\mathrm{c}}=1.6$. After $5$ outer iterations, we obtained the index sets in Figure \ref{fg:index.div.ex2} with $583$ (Figure \ref{fg:ex6.div.index}) and $477$ (Figure \ref{fg:ex6.curl.index}) entries for the divergence- and curl-free fields, respectively.

\begin{figure}
\centering
\subfloat[$\cl{J}(-4,:,:)$]{\label{fg:index.curl.1}\qquad\qquad\qquad\includegraphics[clip,trim={0 0 0 0}]{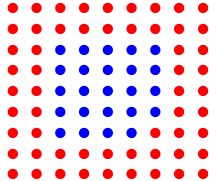}\qquad\qquad\qquad} \hspace{-4cm}
\subfloat[$\cl{J}(-3,:,:)$]{\label{fg:index.curl.2}\qquad\qquad\qquad\includegraphics[clip,trim={0 0 0 0}]{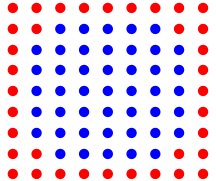}\qquad\qquad\qquad} \hspace{-4cm}
\subfloat[$\cl{J}(-2,:,:)$]{\label{fg:index.curl.3}\qquad\qquad\qquad\includegraphics[clip,trim={0 0 0 0}]{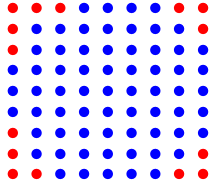}\qquad\qquad\qquad}\\
\subfloat[$\cl{J}(-1,:,:)$]{\label{fg:index.curl.4}\qquad\qquad\qquad\includegraphics[clip,trim={0 0 0 0}]{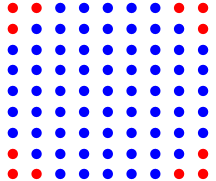}\qquad\qquad\qquad} \hspace{-4cm}
\subfloat[$\cl{J}(0,:,:)$]{\label{fg:index.curl.5}\qquad\qquad\qquad\includegraphics[clip,trim={0 0 0 0}]{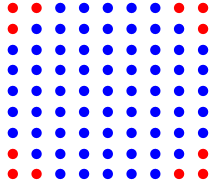}\qquad\qquad\qquad} \hspace{-4cm}
\subfloat[$\cl{J}(1,:,:)$]{\label{fg:index.curl.6}\qquad\qquad\qquad\includegraphics[clip,trim={0 0 0 0}]{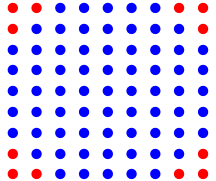}\qquad\qquad\qquad}\\
\subfloat[$\cl{J}(2,:,:)$]{\label{fg:index.curl.7}\qquad\qquad\qquad\includegraphics[clip,trim={0 0 0 0}]{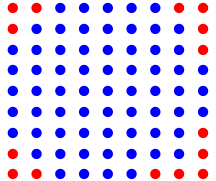}\qquad\qquad\qquad} \hspace{-4cm}
\subfloat[$\cl{J}(3,:,:)$]{\label{fg:index.curl.8}\qquad\qquad\qquad\includegraphics[clip,trim={0 0 0 0}]{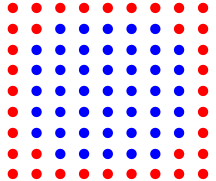}\qquad\qquad\qquad} \hspace{-4cm}
\subfloat[$\cl{J}(4,:,:)$]{\label{fg:index.curl.9}\qquad\qquad\qquad\includegraphics[clip,trim={0 0 0 0}]{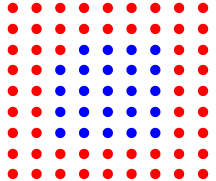}\qquad\qquad\qquad}
\caption{Index set: curl-free field. Panels \protect\subref{fg:index.curl.1}, \protect\subref{fg:index.curl.2}, \protect\subref{fg:index.curl.3}, \protect\subref{fg:index.curl.4}, \protect\subref{fg:index.curl.5}, \protect\subref{fg:index.curl.6}, \protect\subref{fg:index.curl.7}, \protect\subref{fg:index.curl.8}, and \protect\subref{fg:index.curl.9}: $\cl{J}(-4,:,:)$, $\cl{J}(-3,:,:)$, $\cl{J}(-2,:,:)$, $\cl{J}(-1,:,:)$, $\cl{J}(0,:,:)$, $\cl{J}(1,:,:)$, $\cl{J}(2,:,:)$, $\cl{J}(3,:,:)$, and $\cl{J}(4,:,:)$, respectively. Blue dots represent the indices retained in the Fourier construction. Red dots represent indices removed from the Fourier construction.}
\label{fg:index.curl.ex2}
\end{figure}

\begin{figure}
\centering
\subfloat[\PiSL\, field reconstruction]{\label{fg:3d.div.curl}\includegraphics[clip,trim={590 0 590 0},width=0.33\textwidth]{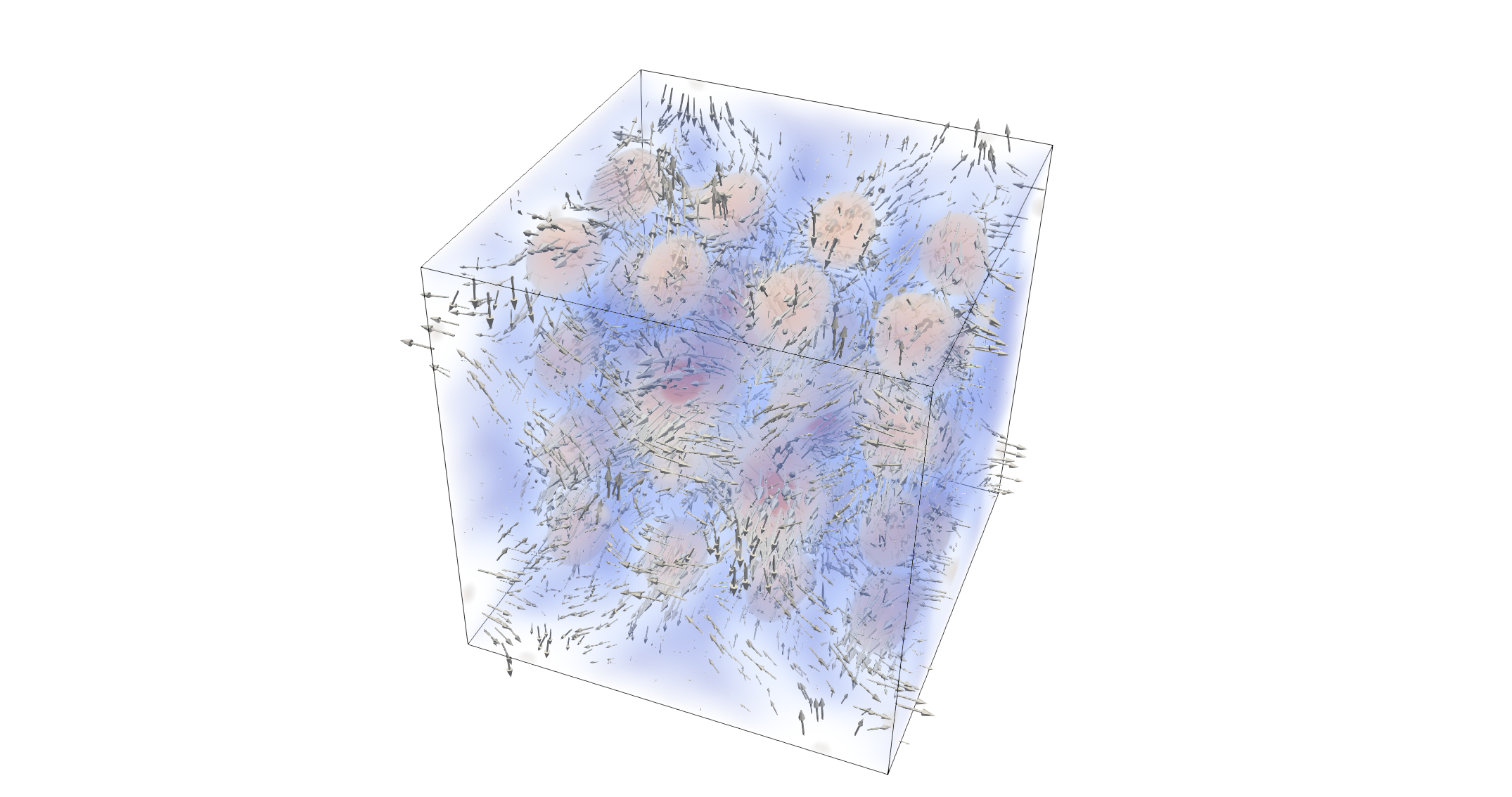}}
\subfloat[\PiSL\, div-free reconstruction]{\label{fg:3d.div}\includegraphics[clip,trim={590 0 590 0},width=0.33\textwidth]{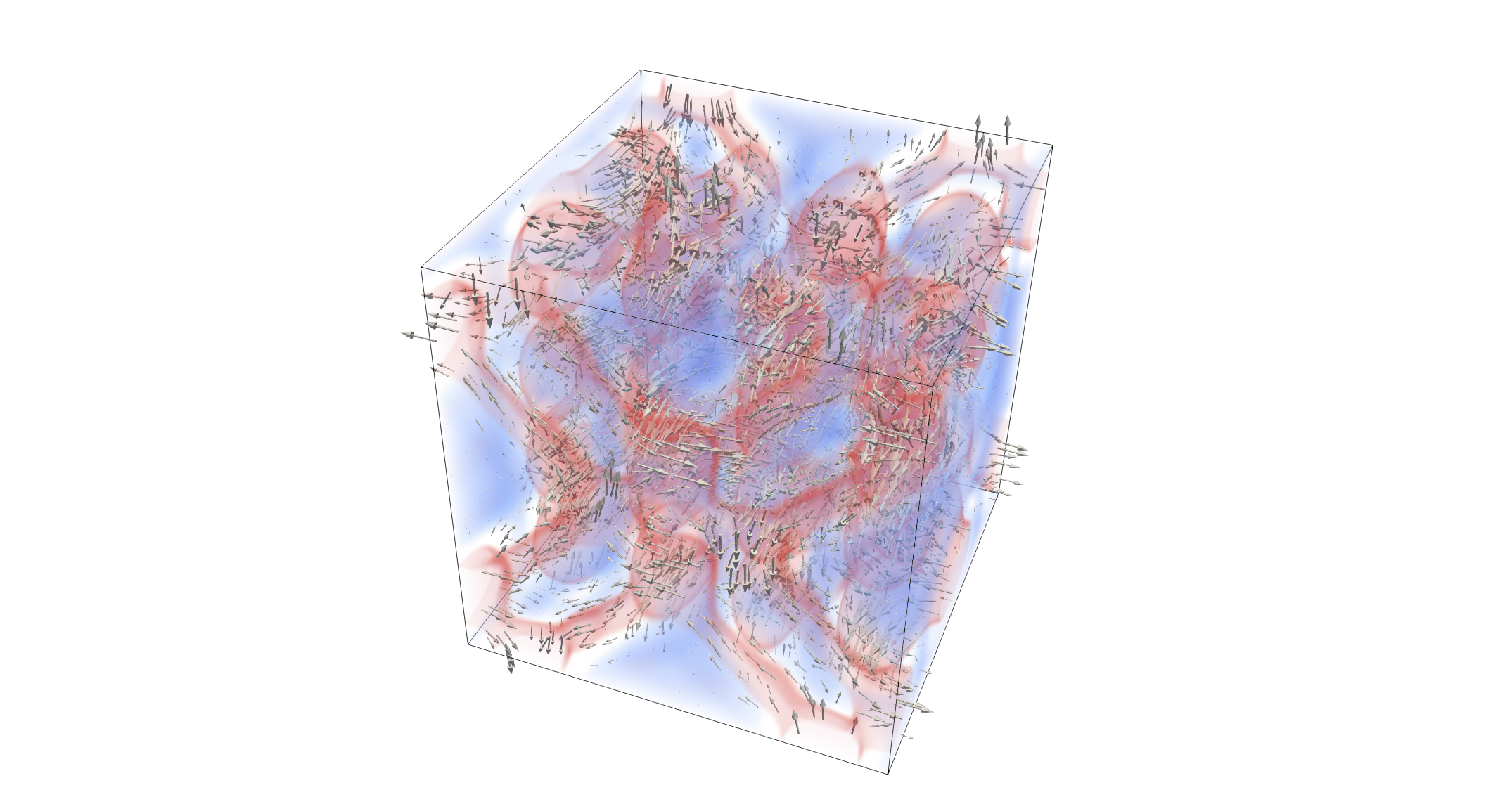}}
\subfloat[\PiSL\, curl-free reconstruction]{\label{fg:3d.curl}\includegraphics[clip,trim={590 0 590 0},width=0.33\textwidth]{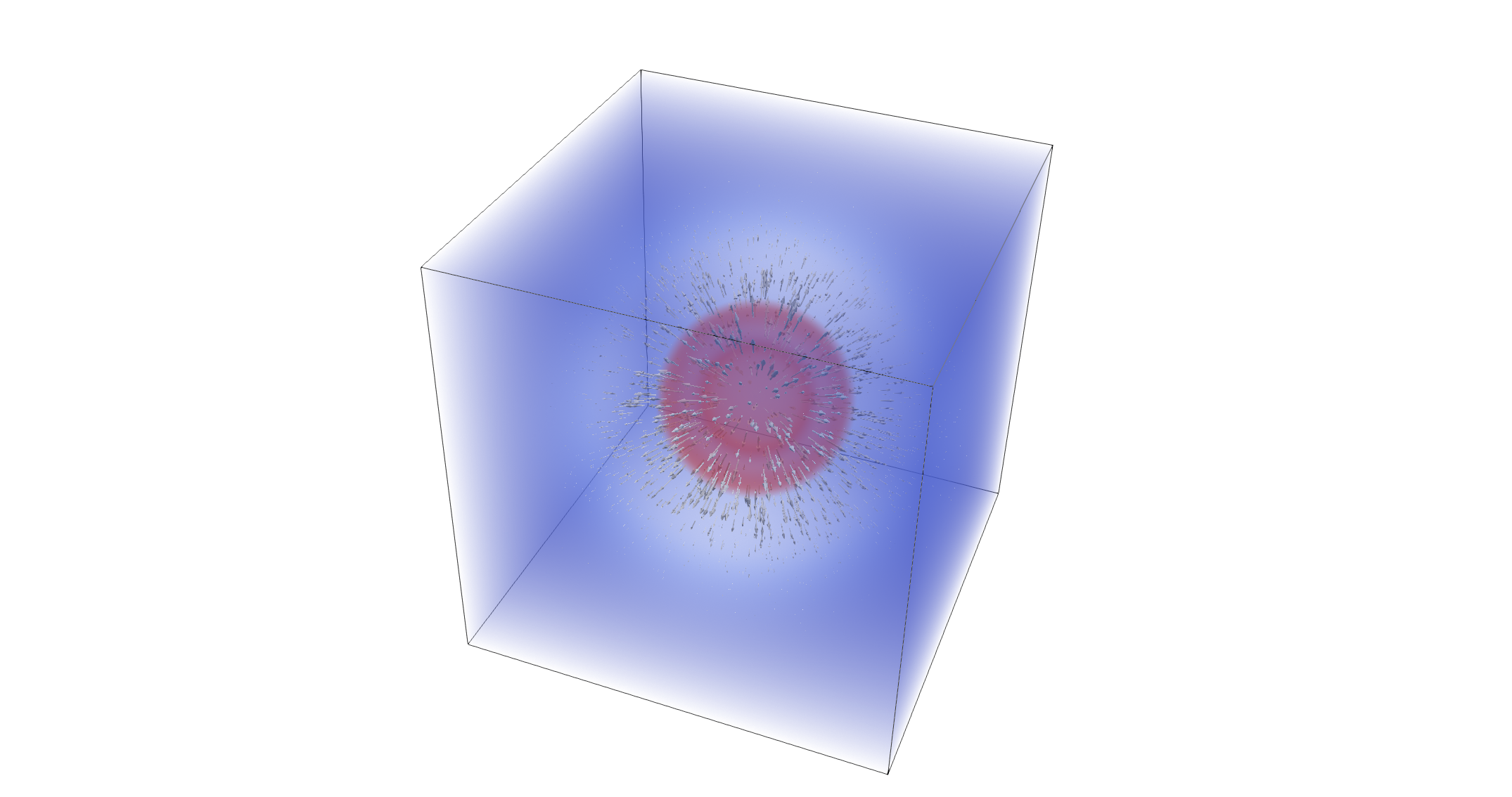}}
\caption{Reconstructed \PiSL\, curl- and divergence-free fields. Panels \protect\subref{fg:3d.div.curl}: $\bsuI+\bssJ$ field, \protect\subref{fg:3d.div}: $\bsuI$ field, and \protect\subref{fg:3d.curl}: $\bssJ$ field.}
\label{fg:fields.ex2}
\end{figure}

\begin{figure}
\centering
\subfloat[\PiSL\, field $L^\infty (e) = 3.2e-2$]{\label{fg:3d.div.curl.err}\includegraphics[clip,trim={560 0 560 0},width=0.33\textwidth]{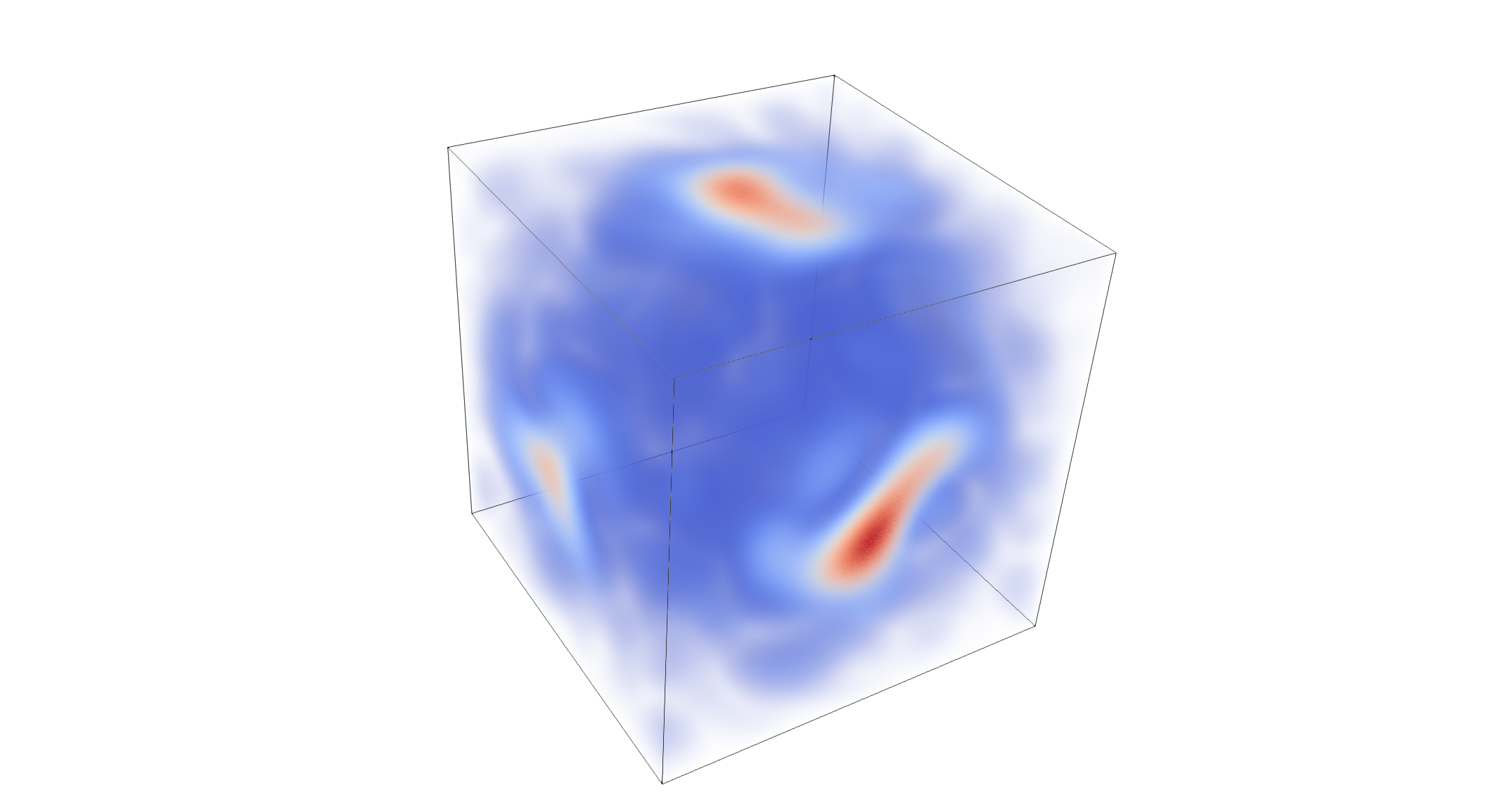}}
\subfloat[\PiSL\, div-free field $L^\infty (e) = 1.1e-2$]{\label{fg:3d.div.err}\includegraphics[clip,trim={560 0 560 0},width=0.33\textwidth]{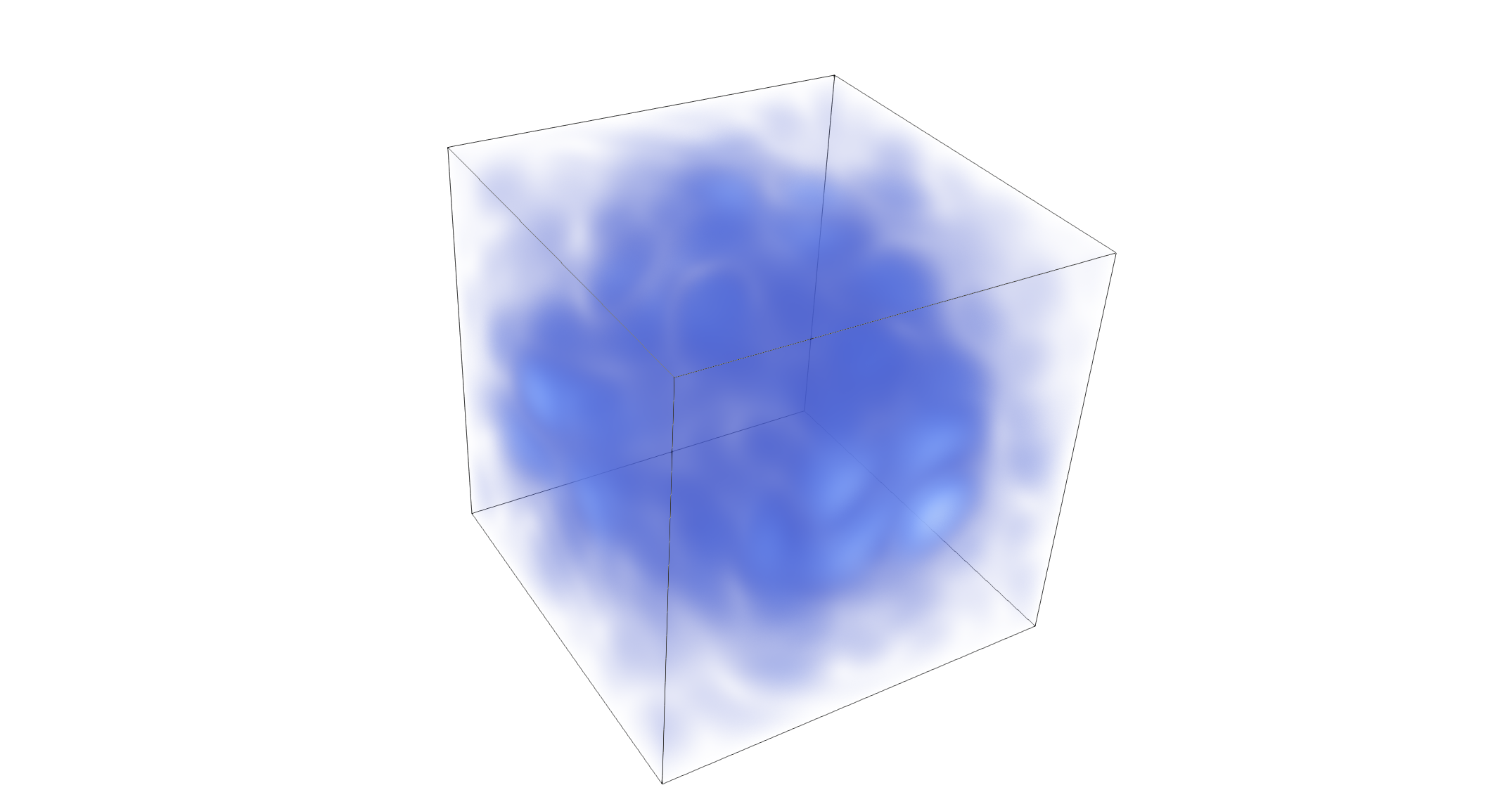}}
\subfloat[\PiSL\, curl-free field $L^\infty (e) = 2.6e-2$]{\label{fg:3d.curl.err}\includegraphics[clip,trim={560 0 560 0},width=0.33\textwidth]{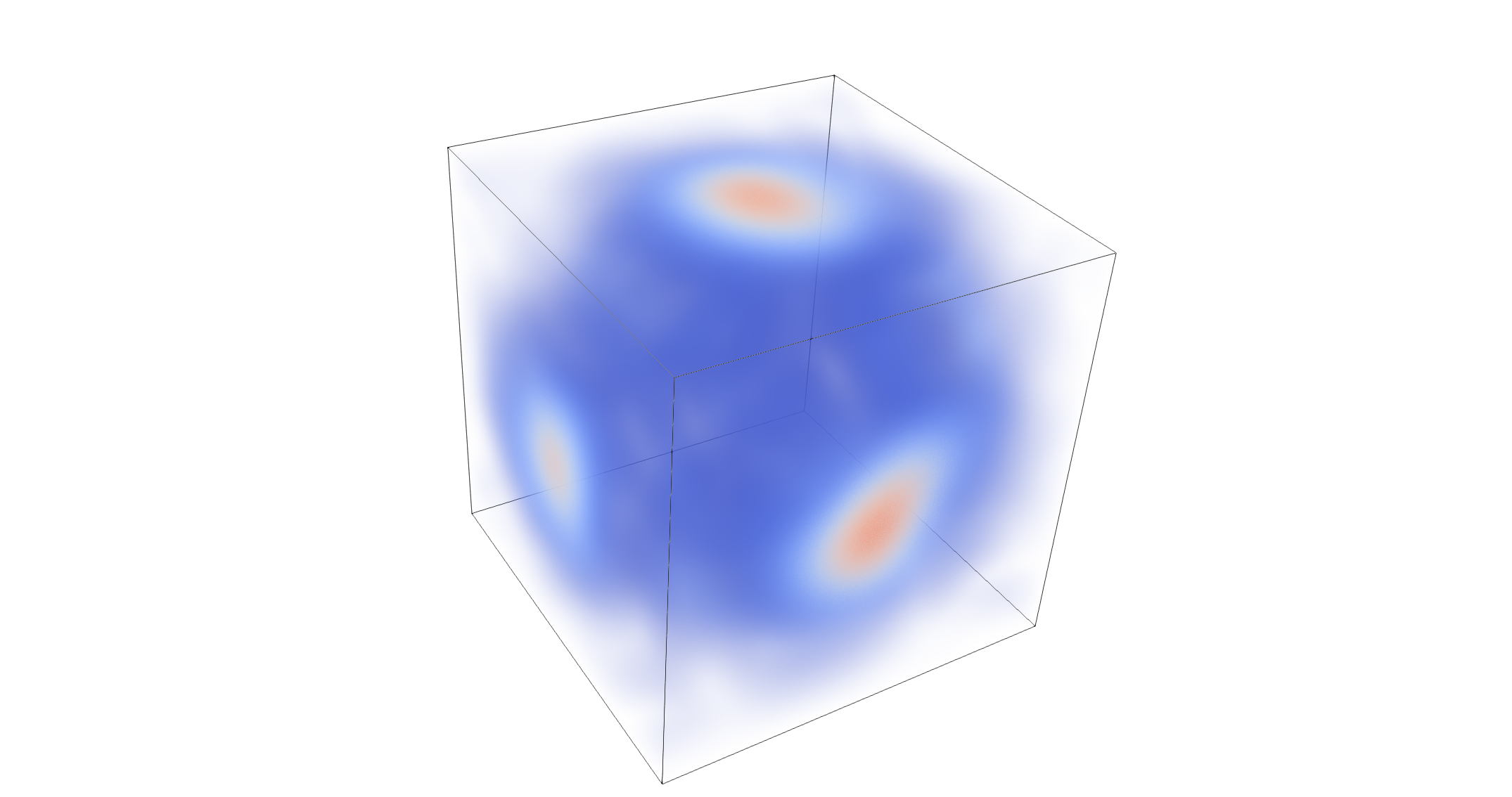}}
\caption{Reconstructed \PiSL\, curl- and divergence-free fields. Panels \protect\subref{fg:3d.div.curl.err}: error in the total field, \protect\subref{fg:3d.div.err}: error in the divergence-free field, and \protect\subref{fg:3d.curl.err}: error in the curl-free field.}
\label{fg:err.ex2}
\end{figure}

Table \ref{tb:ex2} presents the $L^\infty$ and $L^2$ norms of the error $e(\bs{x})$ for the total $\bsuI^{\mathrm{opt}}+\bssJ^{\mathrm{opt}}$, divergence-free $\bsuI$, and curl-free $\bssJ$ fields. As $L^\infty(\|\bs{u}\|) \approx 1.7$, the maximum relative error in any field is less than $2.0\%$. In Figure \ref{fg:fields.ex2}, we present the magnitude of reconstructions from left to right: $\bsuI+\bssJ$, $\bsuI$, and $\bssJ$. Figure \ref{fg:err.ex2} provides the reconstruction errors from left to right, corresponding to $\bsuI+\bssJ$, $\bsuI$, and $\bssJ$ on a scale from $0$ to $0.03$.
\begin{table}[!h]
\caption{$L^\infty$ and $L^2$ norms of the error fields $e$ for the reconstructed total, curl-, and divergence-free fields.}
\begin{tabular}{ c c c }
 $e(\bs{x})$ & $L^2(e(\bs{x}))$ & $L^\infty(e(\bs{x}))$ \\\hline
 $\norm{\bs{u}(\bs{x})-\bsuI^{\mathrm{opt}}(\bs{x})-\bssJ^{\mathrm{opt}}(\bs{x})}{}$ & $0.3742$ & $0.03200$ \\
 $\norm{\bs{u}_{\mathrm{d}}(\bs{x})-\bsuI^{\mathrm{opt}}(\bs{x})}{}$ & $0.2475$ & $0.01126$ \\
 $\norm{\bs{u}_{\mathrm{d}}(\bs{x})-\bsuI^{\mathrm{opt}}(\bs{x})}{}$ & $0.3258$ & $0.02574$ \\
\end{tabular}
\label{tb:ex2}
\end{table}

\FloatBarrier

\subsection{`The storm of the century'}

\begin{figure}
\centering
\includegraphics[clip,trim={0 0 0 0},width=0.65\textwidth]{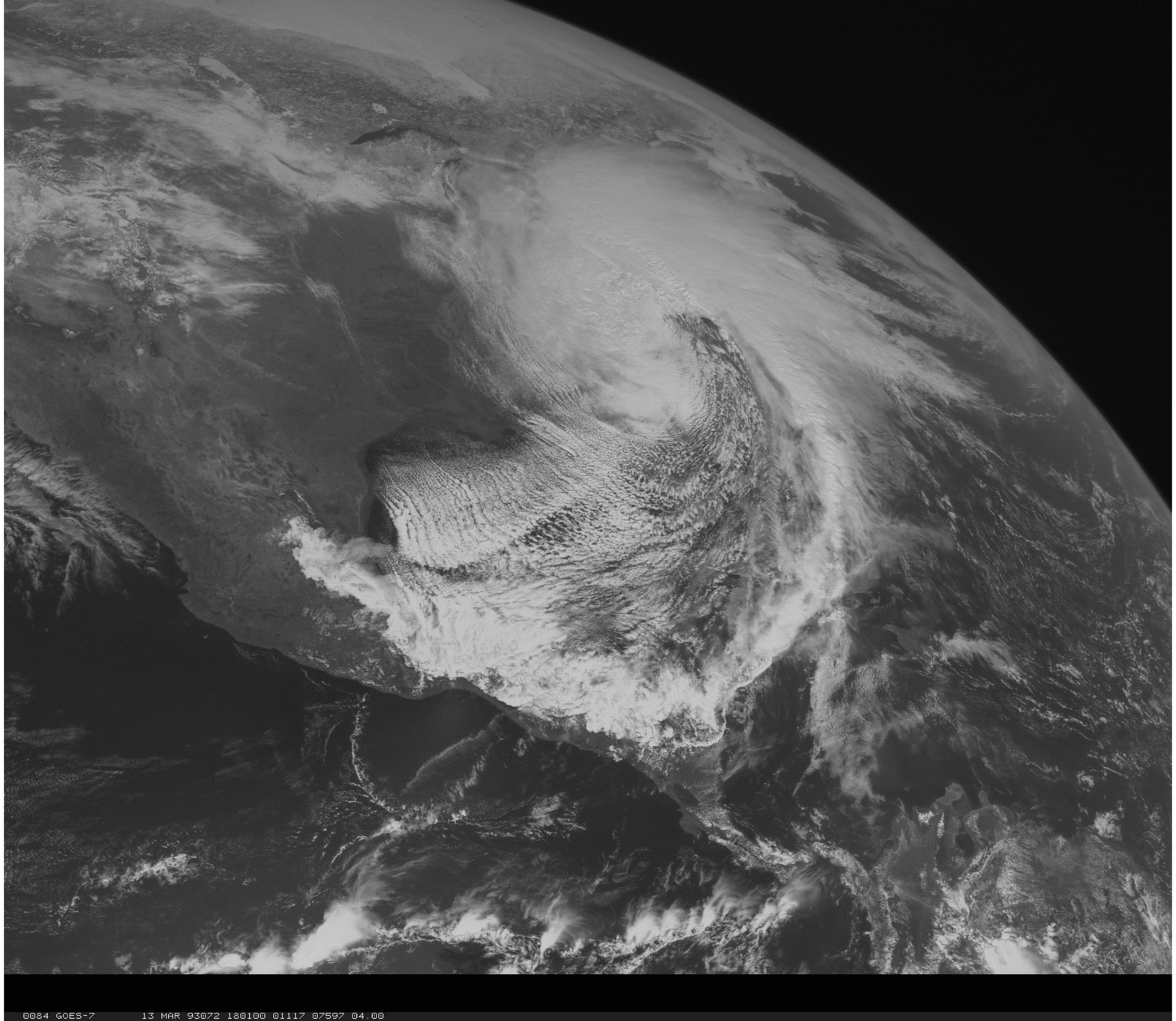}
\caption{March 12--14, 1993 `Storm of the Century' (also known as the 93 Superstorm, No Name Storm, or Great Blizzard of 1993). Picture made available by CIMSS Satellite Blog \url{https://www.ssec.wisc.edu/mcidas/images/goes7_vis_19930313_storm_of_century.jpg}.}
\label{fg:storm}
\end{figure}

The Storm of the Century (also known as the 93 Superstorm, No Name Storm, or Great Blizzard of 1993) occurred from March 12 to 14 in 1993. It was a large cyclonic storm that formed over the Gulf of Mexico. The intensity, massive size, and wide-reaching effects made the storm unique. The storm stretched from Canada to Honduras \cite{Arm16}. Moreover, the storm is one the most significant storms to affect the eastern United States. On March 13, a larger-scale view of Meteosat-3 infrared ($11.5$ $\mu$m) images (Figure \ref{fg:storm}) revealed the vast size of the storm as it moved along the Eastern Seaboard of the US. Some highlights of the storm included snowfall amounts as high as 56 inches at Mount LeConte in Tennessee, wind gusts of $144$ mph in Mount Washington in New Hampshire, a minimum sea level pressure of $28.28$ inches in White Plains in New York, and a post-storm record low temperature of $-24.4$\textcelsius\, in Burlington, Vermont \cite{Sch97}.

Fed by real satellite data, we employ our HHD \PiSL\, framework for hurricane tracking. Therefore, we aim to identify the center of rotation, namely the hurricane eye in the storm of the century, see Figure \ref{fg:storm}. However, the main difficulty arising when dealing with real data is corrupted data. Measurements errors are intrinsic to real data. Therefore, one must avoid overfitting. In the \PiSL\, framework, we control the energy of the higher Fourier modes to assess the quality of the reconstruction and use it as a stopping criterion (see Espath et al. \cite{Esp21}). In this example instead, we limit the energy boundary to $1$\% and $5\%$, for the divergence- and curl-free components, respectively, of the total energy of the approximated solution. This means that we avoid overfitting by not approximating the high freaquences corresponding to $1$\% and $5\%$, for the divergence- and curl-free components, respectively, of the total energy.

\begin{figure}[!h]
\centering
\subfloat[Index set: divergence-free field]{\label{fg:1993.div.index}\qquad\qquad\qquad\includegraphics[]{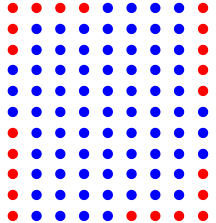}\qquad\qquad\qquad} \hspace{-2cm}
\subfloat[Index set: curl-free field]{\label{fg:1993.curl.index}\qquad\qquad\qquad\includegraphics{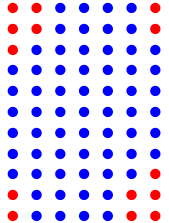}\qquad\qquad\qquad}
\caption{Panels \protect\subref{fg:1993.div.index} and \protect\subref{fg:1993.curl.index}: Index sets of the reconstructed \PiSL\, divergence- and curl-free fields, respectively. Blue dots represent the indices kept in the Fourier construction. Red dots represent indices removed from the Fourier construction.}
\label{fg:index.1993}
\end{figure}

The discrete $L^2$ HHD is carried out using approximately $650$ fixed measurements. We set the residual boundary energy $\varepsilon_{\partial\cl{I}}$ and $\varepsilon_{\partial\cl{J}}$ to $50\%$ and the stopping criteria $\Delta \varepsilon_{\partial\cl{I}}$ and $\Delta \varepsilon_{\partial\cl{J}}$ to $10^{-2}$ and $5 \times 10^{-2}$, respectively. For the fractional Sobolev regularization, we selected $\epsilon_{\mathrm{d}}=\epsilon_{\mathrm{c}}=10^{-3}$ and $k_{\mathrm{d}}=k_{\mathrm{c}}=1.5$. After $9$ outer iterations, we obtained the index sets in Figure \ref{fg:1993.div.index} with $77$ (Figure \ref{fg:ex5.div.index}) and $63$ (Figure \ref{fg:1993.curl.index}) entries for the divergence- and curl-free fields, respectively.

\begin{figure}
\centering
\subfloat[Map]{\label{fg:1993.map}\includegraphics[clip,trim={30 30 30 30},width=0.33\textwidth]{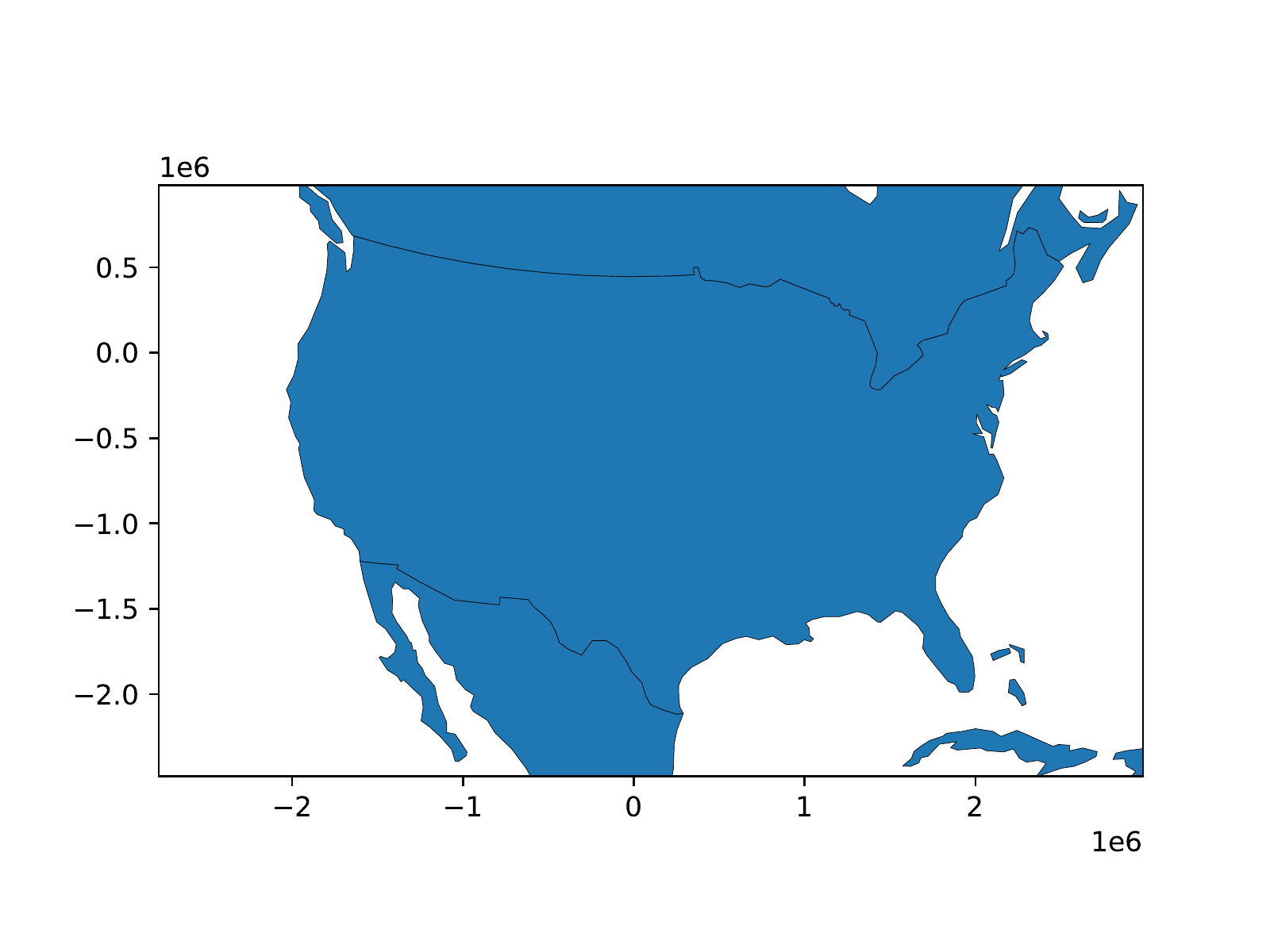}}
\subfloat[Satellite data]{\label{fg:1993.data}\includegraphics[clip,trim={30 30 30 30},width=0.33\textwidth]{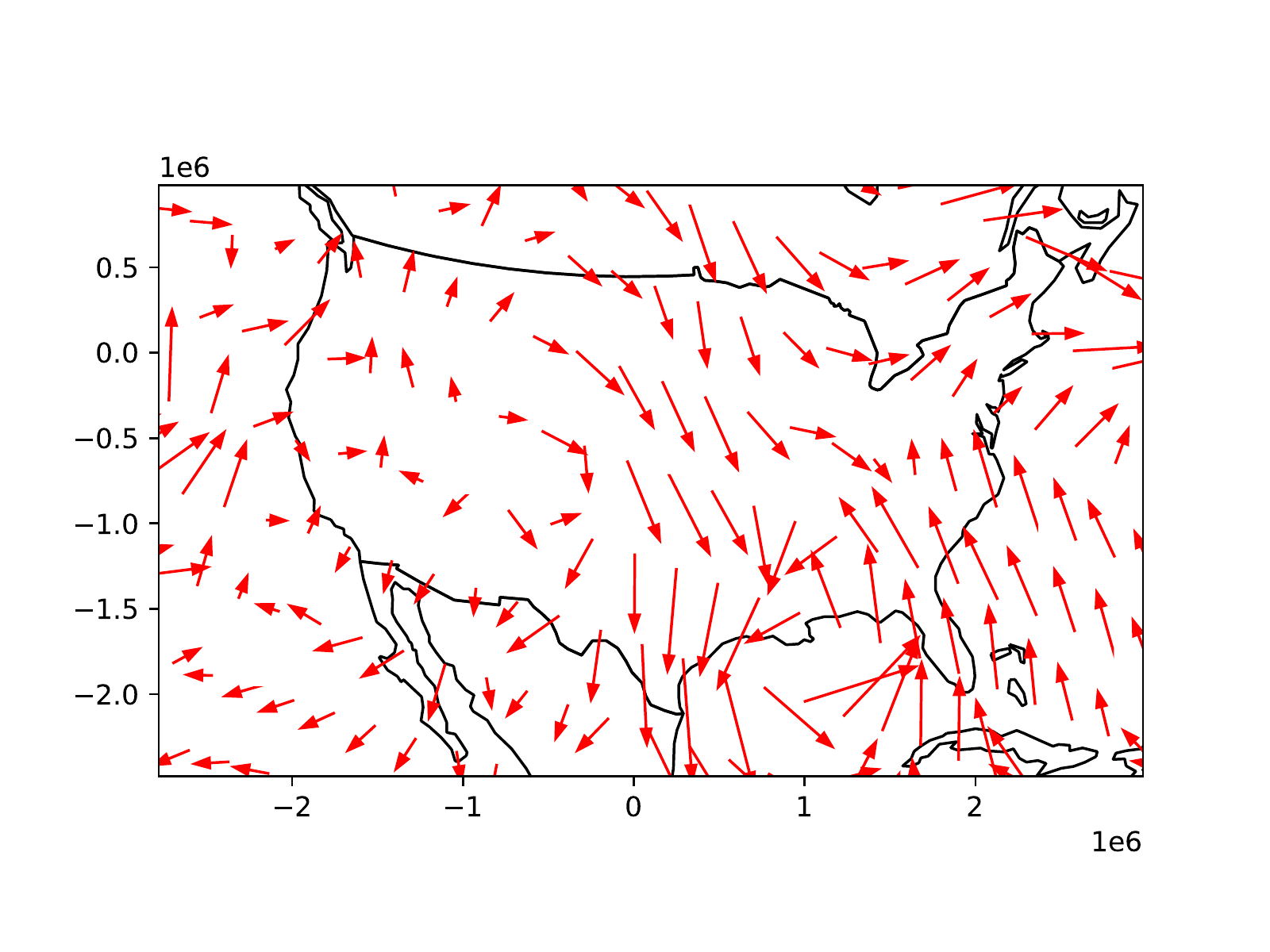}} \\
\subfloat[\PiSL\, wind field reconstrution]{\label{fg:1993.reconstruction}\includegraphics[clip,trim={30 30 30 30},width=0.33\textwidth]{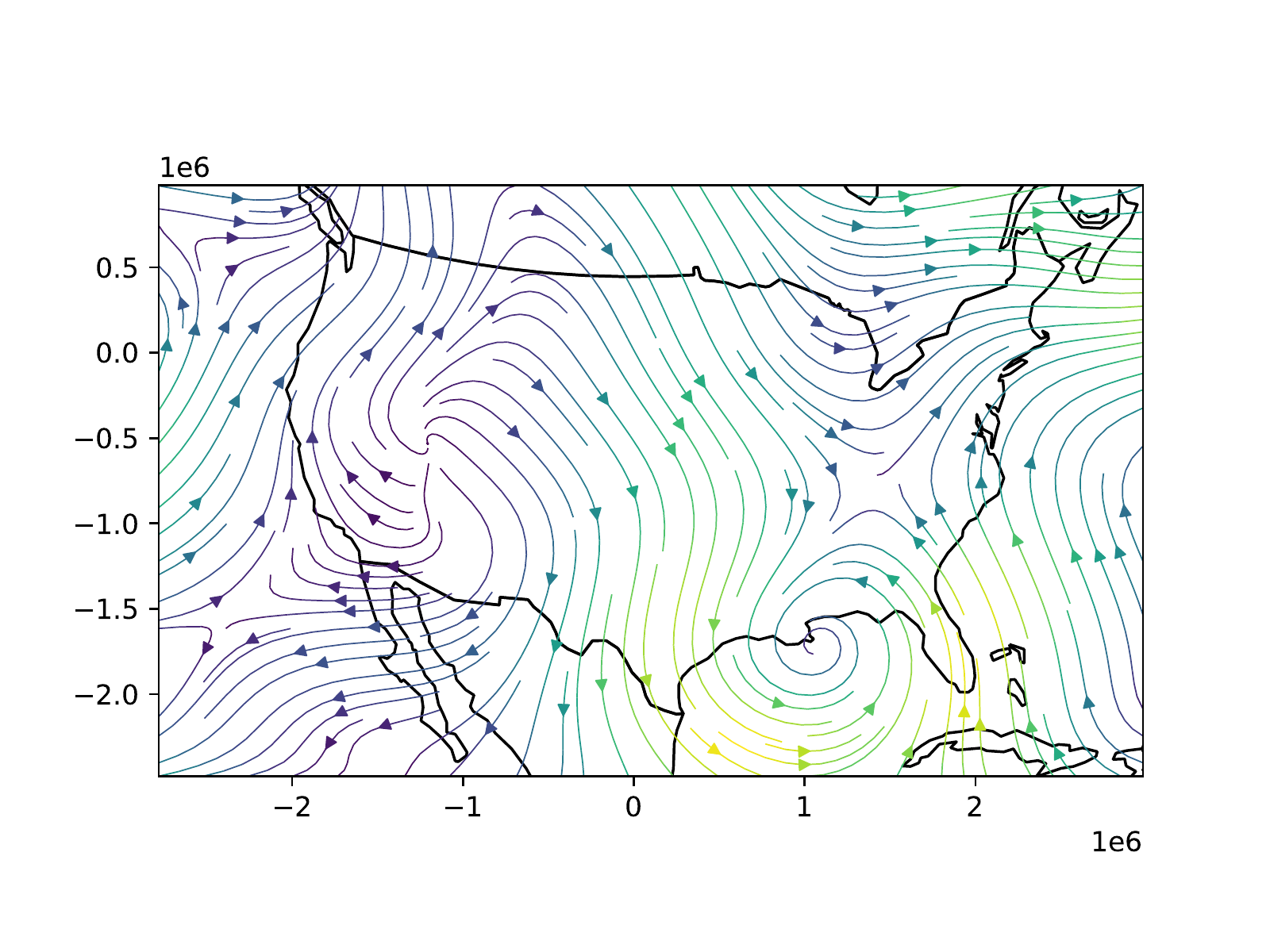}}
\subfloat[\PiSL\, div-free field reconstruction (red dot indicates the hurricane eye)]{\label{fg:1993.div}\includegraphics[clip,trim={30 30 30 30},width=0.33\textwidth]{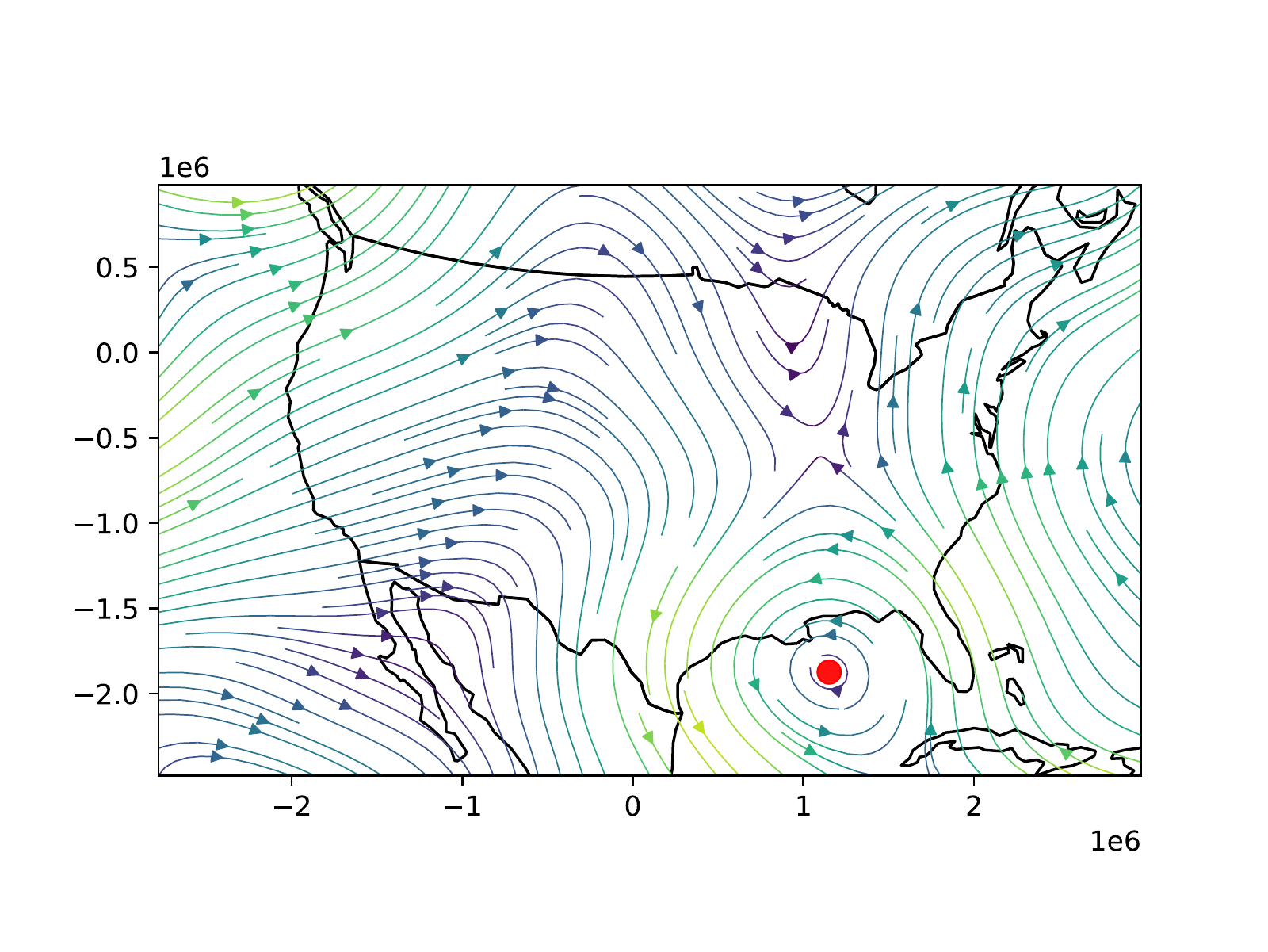}}
\subfloat[\PiSL\, curl-free field reconstruction]{\label{fg:1993.curl}\includegraphics[clip,trim={30 30 30 30},width=0.33\textwidth]{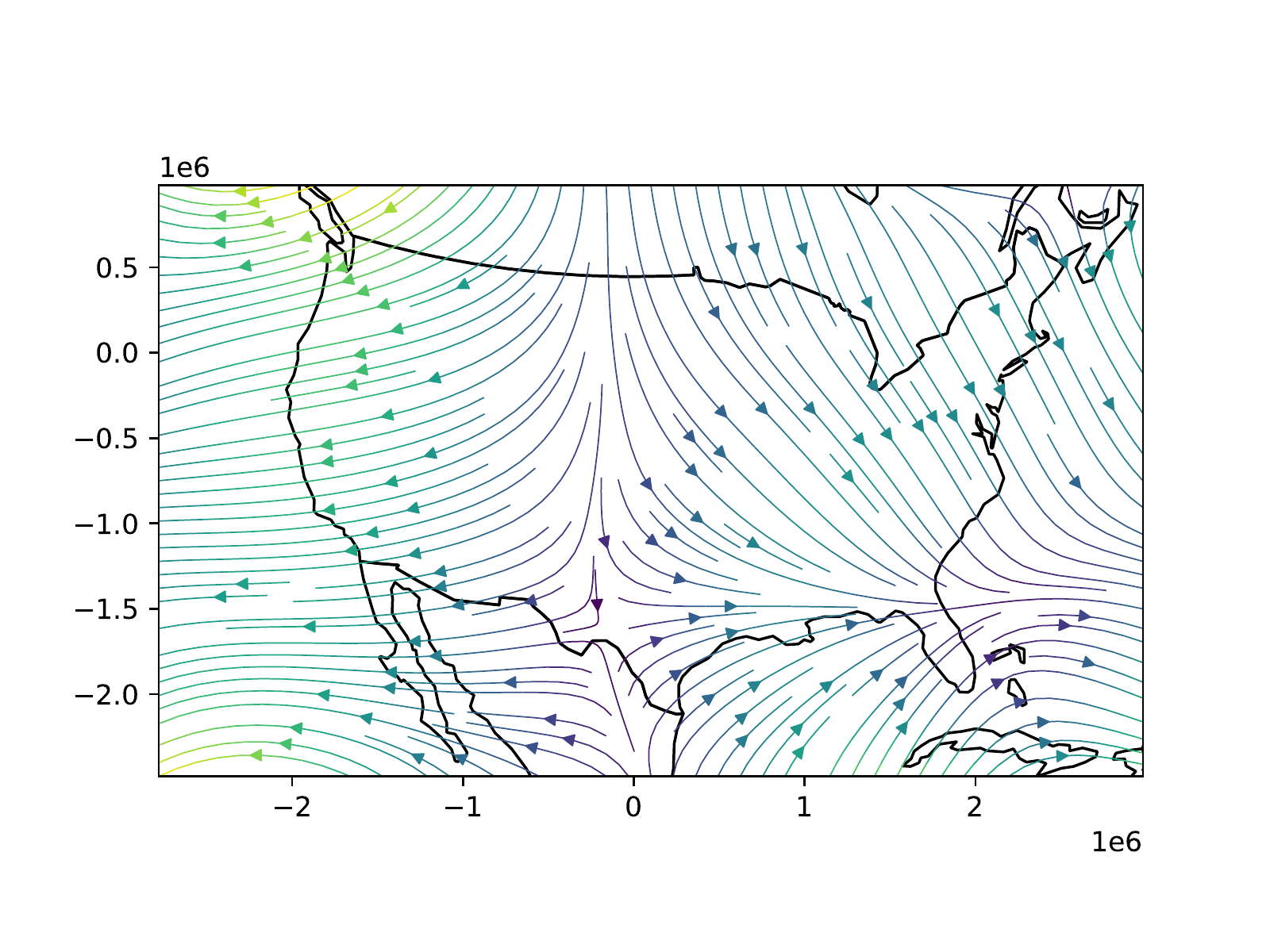}} \\
\subfloat[\PiSL\, vorticity field reconstruction]{\label{fg:1993.vorticity}\includegraphics[clip,trim={30 30 30 30},width=0.33\textwidth]{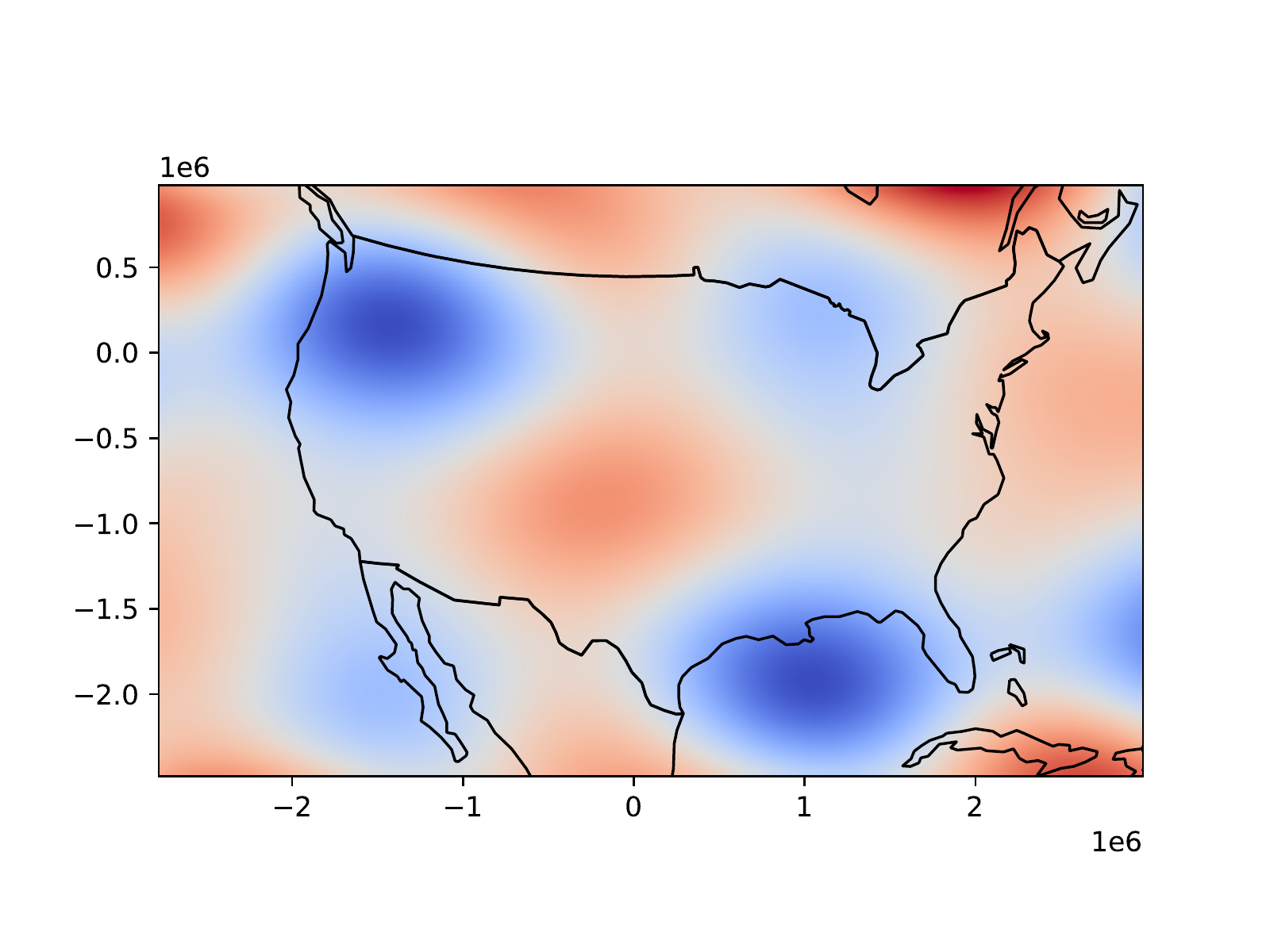}}
\caption{Panels \protect\subref{fg:1993.map}: map, \protect\subref{fg:1993.data}: satellite data, \protect\subref{fg:1993.reconstruction}: \PiSL\, field reconstruction, \protect\subref{fg:1993.div}: \PiSL\, divergence-free field (red dot indicates the hurricane eye), \protect\subref{fg:1993.curl}: \PiSL\, curl-free field, and \protect\subref{fg:1993.vorticity}: \PiSL\, vorticity field. Data from the National Center for Environmental Information (NCEI, formerly NCDC) on a THREDDS server. \url{https://www.ncei.noaa.gov/thredds/catalog/model-narr-a-files/199303/19930313/catalog.html?dataset=model-narr-a-files/199303/19930313/narr-a_221_19930313_0000_000.grb})}
\label{fg:hhd.decomposition.1993}
\end{figure}

In Figure \ref{fg:hhd.decomposition.1993}, we depict the identification of the hurricane eye in the storm of the century in 1993. Figures \ref{fg:1993.map} and \ref{fg:1993.data} present the map and the satellite data used in this discrete HHD \PiSL\, problem, respectively. Figures \ref{fg:1993.div} show the \PiSL\, divergence-free reconstructed component with the red dot indicating the hurricane eye, whereas Figure \ref{fg:1993.curl} presents the \PiSL\, curl-free reconstructed component. Figure \ref{fg:1993.reconstruction} presents the reconstructed \PiSL\, field and Figure \ref{fg:1993.vorticity} presents the \PiSL\, vorticity field. Figure \ref{fg:1993.reconstruction} reveals two main vortical structures, whereas only one vortex is identified in the divergence-free reconstructed component, corresponding to the hurricane eye. The additional `vortex' is an artifact of the two-dimensional data because the satellite preprocess data do not capture the vertical component of the wind velocity. Lastly, in Figure \ref{fg:1993.vorticity}, we present the vorticity of the reconstructed velocity field.

%auto-ignore
\section{Conclusions}
\label{sc:conclusions}

We further develop the Physics-informed Spectral Learning (\PiSL) method introduced by Espath et al. \cite{Esp21} based on a discrete $L^2$ projection to solve the discrete Helmholtz--Hodge decomposition. In this adaptive physical-informed type of statistical learning framework, we adaptively build a sparse Fourier set of basis functions and their coefficients by solving a sequence of minimization problems. The sparse Fourier set of basis functions is augmented greedily in each optimization problem. We regularize our minimization problems with the seminorm of the fractional Sobolev space in a Tikhonov fashion. \PiSL\, method ejoys spectral (exponential) convergence and is powerful enough to reconstruct vector fields from even very sparse data. The reconstruction and identification of the hurricane eye in the storm of the century also depicts the robustness of the discrete $L^2$ HHD using \PiSL.

\section{Acknowledgments}

This work was partially supported by the Flexible Interdisciplinary Research Collaboration Fund at the University of Nottingham Project ID 7466664. This work was partially supported by the KAUST Office of Sponsored Research (OSR) under Award numbers URF$/1/2281-01-01$, URF$/1/2584-01-01$ in the KAUST Competitive Research Grants Program Round 8.

Last but not least, we want to thank Prof. Jesper Oppelstrup for providing us with valuable ideas and constructive comments.

\section{Conflict of interest}

The authors have no conflicts to disclose.

\section{Data Availability}

The data that support the findings of this study are available from the corresponding author upon reasonable request.

%-------------------------------------------------------------------------------%

\appendix
\renewcommand{\theequation}{\thesection.\arabic{equation}}

%auto-ignore
\section{Inner product}
\label{sc:inner.product}

Recalling that
\begin{equation}
\hat{\bs{\alpha}}=(\alpha_1/D_1,\ldots,\alpha_n/D_n)\qquad\text{with}\qquad\bs{\alpha}\in\bb{Z}^n,
\end{equation}
consider
\begin{equation}
\begin{aligned}
\bsxA(\bs{x}) \coloneqq \sum\limits_{\bs{\alpha} \in \cl{A}} \bs{\upsilon}_{\bs{\alpha}} \varphi(\bs{x},\bs{\alpha}) \qquad\text{and}\qquad \bszB(\bs{x}) \coloneqq \sum\limits_{\bs{\alpha} \in \cl{B}} \bs{\varsigma}_{\bs{\alpha}} \varphi(\bs{x},\bs{\alpha}),\\[4pt]
\text{with}\qquad\varphi(\bs{x},\bs{\alpha}) \coloneqq \exp(2\pi\jmath\,\hat{\bs{\alpha}}\cdot\bs{x})\qquad\forall\,\bs{x}\in\Pi_{\beta=1}^n[0,D_\beta]\wedge\bs{\alpha}\in\cl{A},\cl{B},
\end{aligned}
\end{equation}
where $\cl{A}$ and $\cl{B}\subset\bb{Z}^n$ are finite index sets of tuples composed of $n$ integers defining the indices of the basis functions and $\bs{\upsilon}_{\bs{\alpha}},\,\bs{\varsigma}_{\bs{\alpha}}\,\in\bb{C}^n$ are their Fourier coefficients, for all $\bs{\alpha}\in\cl{A}$ and $\bs{\alpha}\in\cl{B}$, respectively.

With the $L^2$ inner product
\begin{equation}
(\bsxA,\bszB)_{L^2(\cl{D})}\coloneqq\dfrac{1}{|\cl{D}|}\int_{\cl{D}}\bsxA\cdot\bszB^\ast\dv,
\end{equation}
and as the bases are orthogonal, that is,
\begin{equation}
(\varphi(\bs{x},\bs{\alpha}),\varphi^\ast(\bs{x},\bs{\beta}))_{L^2(\cl{D})} =
\begin{cases}
0 & \text{if} \, \bs{\alpha} \neq \bs{\beta}, \\[4pt]
1 & \text{if} \, \bs{\alpha} = \bs{\beta}.
\end{cases}
\end{equation}
for $k\in\bb{R}$, and recalling that
\begin{equation}
\Grad\varphi(\bs{x},\bs{\alpha}) = 2 \pi \jmath \mskip3mu \varphi(\bs{x},\bs{\alpha}) \mskip3mu \hat{\bs{\alpha}},
\end{equation}
the $L^2$ inner product between the fractional $k$th gradients of $\bsxA$ and $\bszB$ is given by
\begin{align}\label{eq:parseval.Hk}
(\bsxA,\bszB)_{\mathring{H}^k(\cl{D})} & = \dfrac{1}{|\cl{D}|}\int_{\cl{D}} \Grad^k \bsxA \twovdots (\Grad^k\bszB)^\ast \dv \nonumber\\
&= \dfrac{1}{|\cl{D}|}\int_{\cl{D}} \Bigg(\sum\limits_{\bs{\alpha} \in \cl{A}} \bs{\xi}_{\bs{\alpha}} \otimes \Grad^k\varphi(\bs{x},\bs{\alpha})\Bigg) \twovdots \Bigg(\sum\limits_{\bs{\beta} \in \cl{B}} \bs{\zeta}_{\bs{\beta}}^\ast \otimes \Grad^k\varphi^\ast(\bs{x},\bs{\beta})\Bigg) \dv,\nonumber\\
&= \dfrac{1}{|\cl{D}|}\int_{\cl{D}} \sum\limits_{\substack{\bs{\alpha} \in \cl{A}\\\bs{\beta} \in \cl{B}}} \left(\bs{\xi}_{\bs{\alpha}}\cdot\bs{\zeta}_{\bs{\beta}}^\ast\right) (\Grad^k\varphi(\bs{x},\bs{\alpha})\cdot\Grad^k\varphi^\ast(\bs{x},\bs{\beta})) \dv,\nonumber\\
&= \dfrac{1}{|\cl{D}|}\sum\limits_{\substack{\bs{\alpha} \in \cl{A}\\\bs{\beta} \in \cl{B}}} \bs{\xi}_{\bs{\alpha}}\cdot\bs{\zeta}_{\bs{\beta}}^\ast  \int_{\cl{D}} \Grad^k\varphi(\bs{x},\bs{\alpha})\cdot\Grad^k\varphi^\ast(\bs{x},\bs{\beta}) \dv,\nonumber\\
&= (2\pi)^{2k}\sum\limits_{\bs{\alpha} \in \cl{A}\cap\cl{B}} \left(\bs{\xi}_{\bs{\alpha}}\cdot\bs{\zeta}_{\bs{\alpha}}^\ast \right) (\hat{\bs{\alpha}}\cdot\hat{\bs{\alpha}})^k.
\end{align}

This expression may be specialized to compute the $L^2$ inner product by setting $k=0$, rendering
\begin{equation}\label{eq:parseval.L2}
(\bsxA,\bszB)_{L^2(\cl{D})} = \sum\limits_{\bs{\alpha} \in \cl{A}\cap\cl{B}} \bs{\xi}_{\bs{\alpha}}\cdot\bs{\zeta}_{\bs{\alpha}}^\ast.
\end{equation}

\section{Gradient of the objective function}
\label{sc:grad}

Here, we aim to obtain the stationary point of $\cl{F}$ given in expression \eqref{eq:f} through the optimality condition in expression \eqref{eq:reformulated.optimization} when evaluated at $\bsuI$ and $\bssJ$ with respect to the Fourier coefficients $\bs{\upsilon}_{\bs{\alpha}}$ and $\bs{\varsigma}_{\bs{\alpha}}$, respectively; that is,
\begin{equation}\label{eq:optimality}
2\dfrac{\partial}{\partial\bs{\upsilon}_{\bs{\alpha}}^\ast}\left(\dfrac{1}{P}\sum_{i=1}^P\norm{\bsuI({\bs{x}}_i) + \bssJ({\bs{x}}_i) - {\bs{u}_i}}{}^2 + \epsilon_{\mathrm{d}}\norm{\bsuI}{\mathring{H}_{\mathrm{div}}^{k_{\mathrm{d}}}(\cl{D})}^2 + \epsilon_{\mathrm{d}}\norm{\bssJ}{\mathring{H}_{\mathrm{curl}}^{k_{\mathrm{c}}}(\cl{D})}^2\right)=\bs{0}.
\end{equation}
the interested reader is referred to Espath et al. \cite{Esp21} for Wirtinger calculus.

The first term in the above expression reads
\begin{equation}
2\dfrac{\partial}{\partial\bs{\upsilon}_{\bs{\alpha}}^\ast}(\norm{\bsuI(\bs{x}_i)+\bssJ(\bs{x}_i)-\bs{u}_i)}{}^2) = \dfrac{2}{P}\sum_{i=1}^P \varphi^\ast(\bs{x}_i,\bs{\alpha}) (\bsuI(\bs{x}_i)+\bssJ(\bs{x}_i)-\bs{u}_i), \qquad \forall \bs{\alpha} \in \cl{I},
\end{equation}
and
\begin{equation}
2\dfrac{\partial}{\partial\bs{\varsigma}_{\bs{\alpha}}^\ast}(\norm{\bsuI(\bs{x}_i)+\bssJ(\bs{x}_i)-\bs{u}_i)}{}^2) = \dfrac{2}{P}\sum_{i=1}^P \varphi^\ast(\bs{x}_i,\bs{\alpha}) (\bsuI(\bs{x}_i)+\bssJ(\bs{x}_i)-\bs{u}_i), \qquad \forall \bs{\alpha} \in \cl{J}.
\end{equation}
The last terms, related to the regularizations, read
\begin{equation}
2\dfrac{\partial}{\partial\bs{\upsilon}_{\bs{\alpha}}^\ast}\left(\epsilon_{\mathrm{d}}\norm{\bsuI}{H_{\mathrm{div}}^{k_{\mathrm{d}}}(\cl{D})}^2\right) = 2\epsilon_{\mathrm{d}} (2\pi)^{2k_{\mathrm{d}}} \bs{\upsilon}_{\bs{\alpha}} (\hat{\bs{\alpha}}\cdot\hat{\bs{\alpha}})^{k_{\mathrm{d}}}, \qquad \forall \bs{\alpha} \in \cl{I},
\end{equation}
and
\begin{equation}
2\dfrac{\partial}{\partial\bs{\varsigma}_{\bs{\alpha}}^\ast}\left(\epsilon_{\mathrm{c}}\norm{\bssJ}{H_{\mathrm{curl}}^{k_{\mathrm{c}}}(\cl{D})}^2\right) = 2\epsilon_{\mathrm{c}} (2\pi)^{2k_{\mathrm{c}}} \bs{\upsilon}_{\bs{\alpha}} (\hat{\bs{\alpha}}\cdot\hat{\bs{\alpha}})^{k_{\mathrm{c}}}, \qquad \forall \bs{\alpha} \in \cl{J}.
\end{equation}
From the optimality condition \eqref{eq:optimality}, we finally arrive at
\begin{equation}\label{eq:optimality.explicit.upsilon}
\nabla_{\bs{\upsilon}_{\bs{\alpha}}} \cl{F} = \dfrac{2}{P}\sum_{i=1}^P \varphi^\ast(\bs{x}_i,\bs{\alpha}) (\bsuI(\bs{x}_i)+\bssJ(\bs{x}_i)-\bs{u}_i) + 2\epsilon_{\mathrm{d}} (2\pi)^{2k_{\mathrm{d}}} \bs{\upsilon}_{\bs{\alpha}} (\hat{\bs{\alpha}}\cdot\hat{\bs{\alpha}})^{k_{\mathrm{d}}} = \bs{0}, \qquad \forall \bs{\alpha} \in \cl{I},
\end{equation}
and
\begin{equation}\label{eq:optimality.explicit.varsigma}
\nabla_{\bs{\varsigma}_{\bs{\alpha}}} \cl{F} = \dfrac{2}{P}\sum_{i=1}^P \varphi^\ast(\bs{x}_i,\bs{\alpha}) (\bsuI(\bs{x}_i)+\bssJ(\bs{x}_i)-\bs{u}_i) + 2\epsilon_{\mathrm{c}} (2\pi)^{2k_{\mathrm{c}}} \bs{\upsilon}_{\bs{\alpha}} (\hat{\bs{\alpha}}\cdot\hat{\bs{\alpha}})^{k_{\mathrm{c}}} = \bs{0}, \qquad \forall \bs{\alpha} \in \cl{J}.
\end{equation}

%-------------------------------------------------------------------------------%

\footnotesize

\bibliographystyle{unsrt}
%\bibliography{bib_section-1,bib_section-2,bib_section-3,bib_section-4,bib_section-5}

\end{document}